\documentclass{article}

\usepackage[final,nonatbib]{neurips_2021}

\usepackage[utf8]{inputenc} % allow utf-8 input
\usepackage[T1]{fontenc}    % use 8-bit T1 fonts
\usepackage{hyperref}       % hyperlinks
\usepackage{url}            % simple URL typesetting
\usepackage{booktabs}       % professional-quality tables
\usepackage{amsfonts}       % blackboard math symbols
\usepackage{nicefrac}       % compact symbols for 1/2, etc.
\usepackage{microtype}      % microtypography
\usepackage{xcolor}         % colors

\usepackage{algorithm}
\usepackage{algorithmic}

\usepackage{graphicx}
\usepackage{array}
\usepackage{url}
\usepackage{amssymb}
\usepackage{amsmath}
\usepackage{amsthm}
\usepackage{caption}
\usepackage{subcaption}
\usepackage{wrapfig}
\usepackage{multicol}
\usepackage{blindtext}
\usepackage{enumitem}
\usepackage{tikz}
\usetikzlibrary{tikzmark,calc}

\usepackage{hyperref}
\hypersetup{colorlinks,allcolors=black}

\newtheorem{theorem}{Theorem}[section]

\newtheorem{remark}[theorem]{Remark}

\title{Manifold Topology Divergence: a Framework for  Comparing Data Manifolds}

\author{%
Serguei Barannikov\\
Skolkovo Institute of Science and Technology\\
Moscow, Russia\\
CNRS, IMJ, Paris University, France
\And 
 Ilya Trofimov\\
Skolkovo Institute of Science and Technology\\
Moscow, Russia
\And
Grigorii Sotnikov\\
 Skolkovo Institute of Science and Technology\\
 Moscow, Russia
 \And
Ekaterina Trimbach\\
Moscow Institute of Physics and Technology\\
Moscow, Russia
\And
Alexander Korotin\\
Skolkovo Institute of Science and Technology\\
Moscow, Russia
\And
Alexander Filippov\\
Huawei Noah's Ark Lab
\And
Evgeny Burnaev\\
Skolkovo Institute of Science and Technology,\\ Artificial Intelligence Research Institute (AIRI),\\ Moscow, Russia
}

\begin{document}

\maketitle

\begin{abstract}
%Despite several recent proposals there is still a lack of good practical 
%topological comparison of a pair of distributions supported on low dimensional manifolds. 

We develop a framework for comparing data manifolds, aimed, in particular, towards the evaluation of deep generative models. We describe a novel tool, Cross-Barcode(P,Q), that, given a pair of distributions in a high-dimensional space, tracks multiscale topology spacial discrepancies between manifolds on which the distributions are concentrated. Based on the Cross-Barcode, we introduce the Manifold Topology Divergence score (MTop-Divergence) and apply it to assess the performance of deep generative models in various domains: images, 3D-shapes, time-series, and on different datasets: MNIST, Fashion MNIST, SVHN, CIFAR10, FFHQ, chest X-ray images, market stock data, ShapeNet. We demonstrate that the MTop-Divergence accurately detects various degrees of mode-dropping, intra-mode collapse, mode invention, and image disturbance. Our algorithm scales well (essentially linearly) with the increase of the dimension of the ambient high-dimensional space. It is one of the first TDA-based practical methodologies that can be applied universally to datasets of different sizes and dimensions, including the ones on which the most recent GANs in the visual domain are trained. The proposed method is domain agnostic and does not rely on pre-trained networks.

%We propose a framework for comparing manifolds modeled by data distributions, which is  aimed, in particular, towards the evaluation of deep generative models. We describe a novel tool, Cross-Barcode(P,Q), that, given a pair of distributions in a high-dimensional space, tracks multiscale topology spacial discrepancies between manifolds on which the distributions are concentrated. Based on Cross-Barcode, we introduce the Manifold Topology Divergence score (MTop-Divergence) and apply it to assess the performance of deep generative models in various domains: images, 3D-shapes, time-series, and on different datasets: MNIST, Fashion MNIST, SVHN, CIFAR10, FFHQ, market stock data, ShapeNet. We demonstrate that the MTop-Divergence accurately detects various degrees of mode-dropping, intra-mode collapse, mode invention, and image disturbance. Our algorithm scales well (essentially linearly) with the increase of the dimension of the ambient high-dimensional space.Our methodology can be applied to datasets of different sizes, including the ones on which the recent GANs in the visual domain are trained. (Emphasize that it is one of first TDA application that works for datasets of size up to and dimensions up to . It is one of the first "universal" TDA applicable to various size and different dimensions datasets, including the most recent ones).The proposed method is domain agnostic and does not rely on pretrained networks.
%This method  works  for datasets of arbitrary types  
%where  scores of comparative quality might not necessarily exist. 

\end{abstract}

\section{Introduction}
\label{intr}

Geometric perspective in working with data distributions has been pervasive in machine learning  \cite{belkin2001laplacian,LeCunGeometric2017,goodfellow2016deep, chazal2017introduction,moor2020topological,kim2020pllay}.
%Working with the data distributions from geometric perspective has been pervasively (fruitful) in machine learning
Reconstruction of the data from observing only a subset of its points has made a significant step forward since the invention of Generative Adversarial Networks (GANs) \cite{goodfellow2014generative}. 
Despite the exceptional success that deep generative models achieved, there still exists a longstanding challenge of good assessment of the generated samples quality and diversity \cite{borji2019pros}.

For images, the Fr\'echet Inception Distance (FID) \cite{heusel2017gans} is the most popular GAN evaluation measure. However, FID is limited only to 2D images since it relies on pre-trained on ImageNet ``Inception'' network. FID unrealistically approximates point clouds by Gaussians in embedding space; also, FID is biased \cite{binkowski2018demystifying}.
Surprisingly, FID can't be applied to compare adversarial and non-adversarial generative models since it is overly pessimistic to the latter ones \cite{ravuri2019classification}.

The evaluation of generative models is about comparing two \textit{point clouds}: the true data cloud $P_{\mathrm{data}}$  and the model (generated) cloud $Q_{\mathrm{model}}$. 
In view of the commonly assumed Manifold Hypothesis \cite{belkin2001laplacian,goodfellow2016deep}, we develop a topology-based measure for comparing two manifolds:
%we 
%can think about this problem as comparing two \textit{manifolds}: 
 the true data manifold $M_{\mathrm{data}}$ and the model manifold $M_{\mathrm{model}}$,
%we can think about this problem as comparing two \textit{manifolds} in the same ambient space: the manifold of real  $M_{\mathrm{data}}$ and generated $M_{\mathrm{model}}$ objects.
%We
%approach this problem by means of the topological data analysis (TDA). 
%We develop a topology-based measure for comparing two manifolds,
by analysing samples $P_{\mathrm{data}} \subset M_{\mathrm{data}}$ and $Q_{\mathrm{model}} \subset M_{\mathrm{model}}$.
%from these manifolds.

\textbf{Contribution.} In this work, we make the following contributions:
\begin{enumerate}[topsep=0pt,noitemsep,
partopsep=0pt, parsep=0ex, leftmargin=*]
    \item We introduce a new tool: $\text{Cross-Barcode}(P, Q)$. For a pair of point clouds $P$ and $Q$, the 
%\pagebreak[4] 
    \mbox{$\text{Cross-Barcode}(P, Q)$} records the differences in multiscale topology between two manifolds  approximated by the point clouds;
    \item We propose a new measure for comparing two data manifolds approximated by point clouds: Manifold Topology Divergence (MTop-Div);
    \item We apply the MTop-Div to evaluate performance of GANs in various domains: 2D images, 3D shapes, time-series. We show that the MTop-Div correlates well with domain-specific measures and can be used for model selection. Also it provides insights about evolution of generated data manifold during training;
    \item We have compared the MTop-Div against 7 established evaluation methods: FID, discriminative score, MMD, JSD, 1-coverage, IMD and Geometry score and found that MTop-Div is able to capture subtle differences in data geometry;
    \item We have essentially overcame the known TDA scalability issues and in particular have carried out the MTop-Div calculations on most recent datasets such as FFHQ, with dimensions $D$ up to $10^7$. 
\end{enumerate}
%\begin{enumerate}[topsep=0pt,noitemsep, leftmargin=*]
%    \item We introduce a new tool: $\text{Cross-Barcode}(P, Q)$. For a pair of point clouds $P$ and $Q$, the \mbox{$\text{Cross-Barcode}(P, Q)$} records the differences in multiscale topology between two manifolds  approximated by the point clouds;
%    \item We propose a new measure for comparing two data manifolds approximated by point clouds: Manifold Topology Divergence (MTop-Div);
%    \item We apply the MTop-Div to evaluate performance of GANs in various domains: 2D images, 3D shapes, time-series. We have compared the MTop-Div against 6 established evaluation methods: FID, discriminative score, MMD, JSD, 1-coverage, and Geometry score. We show that the MTop-Div correlates well with domain-specific measures and can be used for model selection. Also it provides insights about evolution of generated data manifold during training;
%    \item  and found that MTop-Div is able to capture subtle differences in data geometry;
%    \item We have essentially overcame the known TDA scalability issues and in particular have carried out the MTop-Div calculations on most recent datasets such as FFHQ, with dimensions $D$ up to $10^7$. 
%\end{enumerate}
The source code is available at \url{https://github.com/IlyaTrofimov/MTopDiv}.

\textbf{Related work.}
%It is well known that the values of the GAN's loss function do not contain a lot of insights about the model.
%only providing information about training dynamics. 
%That is why there exists a longstanding question about assessing the quality of the model by observing its samples. 
%\textbf{Evaluation of GANs} is still an open issue. 
GANs try to recover the true data distribution via a min-max game where two players, typically represented by deep neural networks, called Discriminator and Generator, compete by optimizing the common objective.
Training curves are not informative since generator and discriminator counter each other, thus, the loss values are often meaningless.  
Several other measures were introduced to estimate the quality of GANs.
However, there is no consensus which of them best captures strengths and limitations of the models and can be used for the fair model selection. 
%Typically, measures compare generated samples with training data samples.
%At the first glance, seems to be a good choice, but it 
The likelihood in Parzen window works poorly in high-dimensional spaces and does not correlate with the visual quality of generated samples \cite{theis2015note}.
The Inception Score (IS) \cite{salimans2016improved}
measures both the discriminability and diversity of generated samples. It relies on the pre-trained Inception network and is limited to 2D images domain. While IS correlates with visual image quality, it ignores the true data distribution, doesn't detect mode dropping and is sensitive to image resolution. 
%Various improvements of IS were proposed.
%FID \cite{heusel2017gans}
The Fr\'echet Inception Distance (FID) \cite{heusel2017gans} is a distance between two multivariate Gaussians. These Gaussians 
approximate the features of generated and true data extracted from the last hidden layer of the pretrained Inception network.
Similar variant is KID \cite{binkowski2018demystifying} which computes MMD distance between two distributions of features. The work
\cite{grnarova2019domain} proposed to use the Duality Gap, a notion from the game theory, as a domain agnostic measure for GAN evaluation. 
%The paper \cite{tsitsulin2019shape} 
%Also \cite{grnarova2019domain} described a computationally efficient algorithm for calculating the Duality Gap. 
Another variant of a measure inspired by game theory was proposed in \cite{olsson2018skill}. Conventional precision and recall measures can be also used \cite{lucic2018gans, kynkaanniemi2019improved}. Accurate calculation of precision and recall is limited to simple datasets from low-dimensional manifolds. 
The \emph{Geometry Score} (GScore) \cite{khrulkov2018geometry} is the L2-distance between mean Relative Living Times (RLT) of topological features calculated for the model distribution and the true data distribution. 
%The work \cite{zhou2020evaluating} uses Wasserstein distance between RLTs for this purpose.
%The Geometry Score (GScore) compares topological properties calculated for each of the two manifolds of true and generated data. 
%For both of them, the following procedure is applied: 1) pick the simplicial witness complex based on small subset of landmark points 2) calculate the persistent barcodes 3) for each of the barcodes, calculate the Relative Living Times (RLTs) of H1 homologies; repeat $10^4$ times 4) average RLTs and obtain the Mean Relative Living Time (MRLT). The GScore itself is an L2-distance between MRLTs of true and generated data.
The GScore is domain agnostic, does not involve auxiliary pretrained networks and is not limited to 2D images. However, GScore is not sensitive even to some simple transformations - like constant shift, dilation, or reflection (see our Fig. \ref{fig:rings_H0_sum}, \ref{fig:fives_rotated}).
%in the section \ref{sec:exp-mnist}). 
The barcodes in GScore are calculated approximately, based only on the approximate witness complexes on $64$ landmark points sampled from each distribution.
That's why the procedure is stochastic \linebreak and should be repeated several thousand times for averaging. Thus, the calculation of GScore can be prohibitively long for large datasets. 
%The IS and FID scores are the most popular but they have their known drawbacks and can only be applied to 2d images.
We also refer reader to the comprehensive survey \cite{borji2019pros}.

%when comparing with our score :
%perhaps to itemize, contrary to ours, the g-score:  1) stays the same under any relative displacement of clouds (so in particular cannot be directly used for unsupervised learning, since any shifted cloud would be equally good) 2) is not very robust , as consequence needs averaging over 1000s 1')comparing two barcodes is more tricky than measuring characteristics of single barcode  3) is not monotonic for many types of  disturbances 4)takes often very long time to compute (averaging recommended in the paper N=10000) (days) (prohibitive on several modern datasets for example ffhq) cannot be implemented as differentiable loss function
%because of this gscore is certainly of theoretical importance but practical use is difficult

%
%
%

%\emph{MTop-Divergence}
%Briefly describe what we do and what are 
%\emph{Key differences} compared with g-score and FID
%above + sensitivity to displacement of clouds
%+timing/complexity (no need to use witness which is not available on gpu)
%+possibility to backpropagate (sensitivity to relative position of the clouds is important)->few shots learning?
%+kind of remarkable coincidence of relative positioning with scores measured on pixel representations of clouds and on embedding representations

%T

%+perhaps couple words on using the discriminator loss why it is not good for evaluation of quality
%+ on precision and recall there were other couple of papers: 1904.06991
%Applications of measures: convergence criteria, stopping criteria, hyperparameter tuning, model ranking.

\section{Cross-Barcode and Manifold Topology Divergence }\label{sec:Theor}

\subsection{Multiscale simplicial approximation of manifolds}
%We start from a circle of ideas related with Manifold Hypothesis. 
According to the well-known Manifold Hypothesis \cite{goodfellow2016deep} the support of the data distribution  $\cal{P}_{\mathrm{data}}$ is often concentrated on a low-dimensional manifold $M_{\mathrm{data}}$. We construct a framework for comparing numerically such  distribution $\cal{P}_{\mathrm{data}}$ with a similar distribution $\cal{Q}_{\mathrm{model}}$ concentrated on a  manifold $M_{\mathrm{model}}$. Such distribution $\cal{Q}_{\mathrm{model}}$  is produced, for example,  by a generative deep neural network in one of applications' scenarios. 
The immediate difficulty here is that the manifold $M_{\mathrm{data}}$ is unknown and is described only through  discrete sets of samples from the distribution $\cal{P}_{\mathrm{data}}$. One standard approach to resolve this difficulty is to approximate the manifold $M_{\mathrm{data}}$  by simplices with vertices given by the sampled points. The simplices approximating the manifold are picked based on  proximity information given by the pairwise distances between sampled points \cite{belkin2001laplacian,niyogi2008finding}. The standard approach is to fix a threshold $r>0$ and to take the simplices whose edges do not exceed the threshold $r$.  The choice of threshold is essential here since if it is too small, then only the initial points, i.e., separated from each other 0-dimensional simplices, are allowed. And if the threshold is too large, then all possible simplices with sampled points as vertices are included and their union is simply the big blob representing the convex hull of the sampled points. Instead of trying to guess the right value of the threshold, the standard recent approach is to study all thresholds at once. This can be achieved thanks to the mathematical tool, called barcode \cite{B94,chazal2017introduction}, that quantifies the evolution of topological features over multiple scales. For each value of $r$ the barcode describes the topology, namely the numbers of holes or voids of different dimensions, of the union of all simplices included up to the threshold $r$.  

\subsection{Measuring the differences in simplicial approximation of two manifolds }
However, to estimate numerically the degree of similarity between the manifolds $M_{\mathrm{model}}, M_{\mathrm{data}}\subset {\mathbb{R}}^D$, it is important not just to know the numbers of topological features across different scales for simplicial complexes approximating $M_{\mathrm{model}}, M_{\mathrm{data}}$ , but to be able to verify  that the similar topological features are located  at similar places and appear at similar scales.

\begin{wrapfigure}{r}{0.44\textwidth}
\centering 
\vskip-0.2in
 \includegraphics[width=0.14\textwidth]{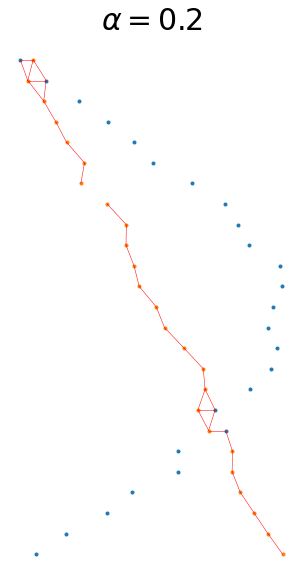} 
 \includegraphics[width=0.14\textwidth]{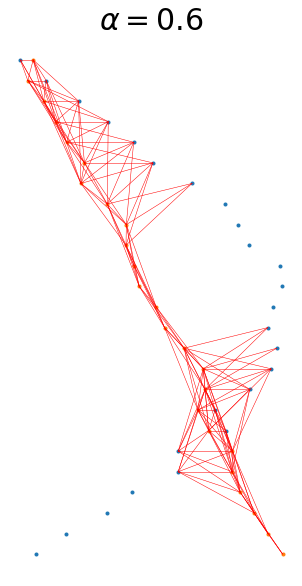}
 \includegraphics[width=0.14\textwidth]{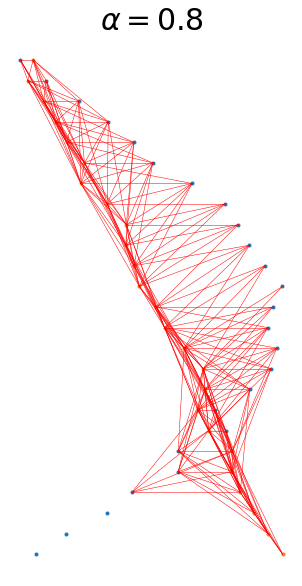}
  \caption{ Edges(red)  connecting $P-$points(red) with $Q-$points(blue),  and also $P-$points between them, are added for three thresholds: $\alpha=0.2, 0.4,0.6$}
 \label{fig:edges}
\vskip-0.15in
\end{wrapfigure}

Our method measures the differences in the simplicial approximation of the two manifolds, represented by samples $P$ and $Q$, by  constructing  sets of simplices,  describing discrepancies  between the two manifolds. To construct these sets of simplices 
 we take the edges connecting $P-$points with $Q-$points,  and also $P-$points between them,  ordered by their length, and start adding these edges one by one, beginning from the smallest edge and gradually increasing the threshold, see Figure \ref{fig:edges}.  We add also  the triangles and $k-$simplices at the threshold  when all their edges have been added. It is assumed that all edges between $Q-$points were already in the initial set. We track in this process the births and the deaths of topological features, where the topological features are allowed here to have boundaries on any simplices formed by $Q-$points. The longer the lifespan of the topological feature across the change of threshold the bigger the described by this feature discrepancy between the two manifolds.

%\begin{figure}[h!]
%\centering 
% \includegraphics[width=0.31\columnwidth]{img/disks/manifolds/alpha20-2.png} 
% \includegraphics[width=0.31\columnwidth]{img/disks/manifolds/alpha40-2.png}
% \includegraphics[width=0.31\columnwidth]{img/disks/manifolds/alpha60-2.png}
%  \caption{ Edges(red)  connecting $P-$points(yellow) with $Q-$points(blue),  and also $P-$points between them, are added for three thresholds: $\alpha=0.2$, $\alpha=0.4$, $\alpha=0.6$}
 %\label{fig:edges}
%\end{figure}

 \textbf{Homology} is a tool that  permits to single out topological features that are similar, and to decompose any topological feature into a sum of  basic  topological features. More specifically, 
 %we use the homology of a pair. I
 in our case,   a $k-$cycle is a collection of $k-$simplices formed by $P-$ and $Q-$ points, such that their boundaries cancel each other, perhaps except for the boundary simplices formed only by $Q-$points.  For example,  a cycle of dimension $k=1$ corresponds to a path connecting a pair of $Q-$points and consisting of edges passing through a set of $P-$points. A cycle which is a boundary of a set of  $(k+1)-$simplices is considered trivial. Two cycles are topologically equivalent if they differ by a boundary, and by collection of simplices formed only by $Q-$points. A union of cycles is again a cycle. Each cycle can be represented by a vector in the vector space where each simplex corresponds to a generator.  In practice, the vector space over $\{0,1\}$ is used most often. The union of cycles corresponds to the sum of vectors. The homology vector space $H_k$ is defined as the factor of the vector space of all $k-$cycles modulo the vector space of  boundaries and cycles consisting of simplices formed only by $Q-$points. A set of vectors forming a basis in this factor-space corresponds to a set of basic topological features, so that any other topological feature  is equivalent to a sum of some basic features.  
 
 The homology are also  defined for manifolds and for arbitrary topological spaces. This  definition is technical and we have to omit it due to limited space, and to refer to e.g. \cite{hatcher2005algebraic,moraleda2019computational} for details. The relevant properties for us are the following. For each topological space $X$ the vector spaces $H_k(X)$, $k=0,1\dots$, are defined. The dimension of the vector space $H_k$  equals to the number of independent $k-$dimensional topological features (holes,voids etc). An inclusion    $Y\subset X$ induces a natural map $H_k(Y)\rightarrow H_k(X) $

In terms of homology, we would like to verify that not just the dimensions of homology groups $H_*(M_{\mathrm{model}})$ and $H_*(M_{\mathrm{data}})$ are the same but that more importantly the natural maps: 
\begin{align}
  \varphi_r:  H_*(M_{\mathrm{model}}\cap M_{\mathrm{data}})&\rightarrow H_*(M_{\mathrm{model}}) \\
     \varphi_p: H_*(M_{\mathrm{model}}\cap M_{\mathrm{data}})&\rightarrow H_*(M_{\mathrm{data}})
\end{align}
induced by the embeddings are as close as possible to isomorphisms. 
 The homology of a pair is the tool that measures how far such maps are from isomorphisms. Given a pair of topological spaces $Y\subset X$, the homology of a pair $H_*(X,Y)$
counts the number of independent topological features in $X$ that cannot be deformed to a topological feauture in $Y$ plus independent topological features in $Y$ that, after the embedding to $X$, become deformable to a point.  An equivalent description,  the homology of a pair $H_*(X,Y)$  counts the number of independent topological features in the factor-space $X/Y$, where all points of $Y$ are contracted to a single point. The important fact for us is that the map, induced by the embedding, $H_*(Y)\rightarrow H_*(X)$ is an isomorphism if and only if the homology of the pair  $H_*(X,Y)$ are trivial. 
Moreover the  embedding of  simple simplicial complexes $Y\subset X$  is an equivalence in homotopy category, if and only if the homology of the pair $H_*(X,Y)$ are trivial \cite{whitehead2012elements}.

To define the counterpart of this construction for manifolds represented by point clouds, we employ the following strategy. Firstly, we replace the pair $(M_{\mathrm{model}}\cap M_{\mathrm{data}})\subset M_{\mathrm{model}}$ by the equivalent pair $M_{\mathrm{model}}\subset (M_{\mathrm{data}}\cup M_{\mathrm{model}})$ with the same factor-space. Then, we represent   $(M_{\mathrm{data}}\cup M_{\mathrm{model}})$ by the union of point clouds $P\cup Q$, where the  point clouds $P$, $Q$ are sampled from the distributions $\cal{P}_{\mathrm{data}}$, $\cal{Q}_{\mathrm{model}}$.  Our principal claim here is that  taking topologically the quotient of $(M_{\mathrm{data}}\cup M_{\mathrm{model}})$ by  $M_{\mathrm{model}}$ is equivalent in the framework of multiscale analysis of topological features to the following operation on the matrix $m_{P\cup Q}$ of pairwise distances of the cloud $P\cup Q$: \textbf{we  set to zero all pairwise distances within the subcloud} ${\bf{Q}} \subset(P\cup Q)$.  
%Since given the  point clouds $P$, $Q$ sampled from the distributions $\cal{P}$, $\cal{Q}$
%the This pair is immediately  It is possible to construct the cloud that represents an intersection of two clouds, we replace 

\subsection{Cross-Barcode\,(P,Q)} \label{}

Let $P=\{p_i\}$, $Q=\{q_j\}$, $p_i, q_j\in {\mathbb{R}}^D$ are two point clouds sampled from two distributions $\cal{P}$, $\cal{Q}$. 
To define $\text{Cross-Barcode}(P,Q)$ we construct first the following filtered simplicial complex. Let $(\Gamma_{P\cup Q},m_{(P\cup Q)/Q})$ be the weighted graph with the distance-like weights on edges defined as the complete graph on the union of point clouds $P\cup Q$ with the distance matrix given by  the pairwise  distance in $\mathbb{R}^D$ for the pairs of points $(p_i,p_j)$ or $(p_i,q_j)$ and with all pairwise distances within the cloud $Q$ that we set to zero. 
Our filtered simplicial complex is the Vietoris-Rips complex of $(\Gamma_{P\cup Q},m_{(P\cup Q)/Q})$. 

Recall that given such a graph $\Gamma$ with matrix $m$ of pairwise distances between vertices  and a
parameter $\alpha>0$, the Vietoris-Rips complex $R_\alpha(\Gamma,m)$ is the abstract simplicial complex with simplices that
correspond to the non-empty subsets of vertices of $\Gamma$  whose pairwise distances are less than $\alpha$ as measured by $m$. Increasing parameter $\alpha$ adds more simplices and this gives a nested family of collections of simplices know as filtered simplicial complex. 
%Such nested family of simplices is known as filtered simplicial complex. 
Recall that a \textbf{simplicial complex} is described by a set of vertices $V=\{v_1,\ldots,v_N\}$, and a collection of $k-$simplices $S$, i.e. $(k+1)-$elements subsets of the set of vertices $V$, $k\geq 0$. The set of simplices $S$ should satisfy the condition that for each simplex $s\in S$ all the $(k-1)$-simplices obtained by the deletion of a vertex from the subset of vertices of $s$ belong also to $S$. The \textbf{filtered simplicial complexes} is the family of simplicial complexes $S_\alpha$ with nested collections of simplices: for $\alpha_1<\alpha_2$ all simplices of $S_{\alpha_1}$ are also in $S_{\alpha_2}$. 

At the initial moment, $\alpha=0$, the simplicial complex $R_\alpha(\Gamma_{P\cup Q},m_{(P\cup Q)/Q})$ has trivial homology $H_k$ for all $k>0$ since it contains all simplices formed by $Q-$points. The dimension of the $0-$th homology  equals at $\alpha=0$  to the number of $P-$points, since no edge between them or between a $P-$point and a $Q-$point is added at the beginning.  As we increase  $\alpha$, some cycles, holes or voids  appear in our complex $R_{\alpha}$. Then, some combinations of these cycles disappear.  The \textbf{persistent homology} principal theorem \cite{B94,zomorodian2001computing} implies that it is possible to choose the set of generators in the homology of  filtered complexes $H_k(R_{\alpha})$ across all the scales $\alpha$ such that each generator appears at its specific "birth" time and disappears at its specific "death" time. These sets of ``birth" and ``deaths"  of topological features in 
$R_{\alpha}$ are registered in \textbf{Barcode} of the filtered complex.
%, \ref{fig:gaussians_barcodes}.  

\textbf{Definition.} The $\text{Cross-Barcode}_{i}(P,Q)$ is the set of intervals recording the ``births" and  ``deaths" times  of $i-$dimensional topological features in the filtered simplicial complex $R_\alpha(\Gamma_{P\cup Q},m_{(P\cup Q)/Q})$. 

Examples of $\text{Cross-Barcode}_{i}(P,Q)$ are shown on Fig. \ref{fig:rings_H0_sum}, \ref{fig:visualization_barcodes}, \ref{fig:mode-dropping}, \ref{fig:stylegan_barcodes}, \ref{fig:gaussians_barcodes}. Topological features with longer ``lifespan" are considered essential.
%, and topological features with a short ``lifespan" are considered less essential.
The topological features with ``birth"=``death" are  trivial by definition and do not appear in $\text{Cross-Barcode}_{*}(P,Q)$.

%sets of simplices $S_{\alpha}$ in the family of simplicial complexes $R_\alpha(\Gamma,m)$  are nested: 
%this defines a filtered simplicial complex, that is, a nested family of simplicial complexes.  
% is an increasing  

%Example (few points) Fig

%As we increase the parameter $\alpha$ families of simplicial complexes certain holes or voids are born and others die 

%\textbf{Barcodes}
%Given a filtered simplicial complex the persistence barcode is defined via the reduction of the matrices of differentials. 
% filtered simplicial complex obtained by forming a simplex for each set of vertices of the graph that are pairwise connected by edges of length less than a given threshold. 

%\emph{Remark.} This filtered simplicial complex is different from a factor complex obtained from factoring the Vietoris-Rips complex on $Q-$vertices .

%The manifolds 

%One source of inspiration for  our construction comes from a version of  Whitehead theorem in algebraic topology.  So if two manifolds coincide then An additional difficulty  in our case  is that the manifolds are discretized.  

\subsection{Basic properties of   $\text{Cross-Barcode}_{*}$(P,Q)}\label{basicprop}
\textbf{Proposition 1.} Here is a list of basic properties of  $\text{Cross-Barcode}_{*} (P,Q)$:

\begin{itemize}[topsep=0pt,noitemsep,nolistsep, partopsep=0pt, parsep=0ex, leftmargin=*]
  \item if the two clouds coincide then $\text{Cross-Barcode}_{*} (P,P)=\varnothing$;
  %\item The same is true for any subset $P'\suset P$, including the empty subset:
   %$\text{Cross-Barcode}_{*} (P',P)=\varnothing$, factoring a subset by the bigger set is obviously 
  \item for $Q=\varnothing$, $\text{Cross-Barcode}_{*} (P,\varnothing)=\text{Barcode}_{*} (P)$, the barcode of the single point cloud $P$ itself; 
  %factoring by the empty cloud changes nothing and
  \item the norm of  $\text{Cross-Barcode}_{i} (P,Q)$, $i\geq 0$, is bounded from above by the Hausdorff distance 
  %between $P$ and $Q$ 
  \begin{equation}\label{eq:hausd}
      \left\Vert \text{Cross-Barcode}_{i} \left(P,Q\right)\right\Vert_B\leq d_H(P,Q).
  \end{equation}
The proof is given in appendix.

\end{itemize}
%$\text{Cross-Barcode}_{*} (P,P)=0$

%Triples of clouds, chain rule, exact sequences

%First properties (shift, rotations of both clouds, affine transformation of both clouds)

%Stability. The  is useful for applications because it is stable with respect
%to small perturbations in the input data.

%The first stability theorem for PH, proven in , asserts that, under favorable conditions, in the pipeline in Figure  is 1-Lipschitz with respect to suitable distance
%functions on filtered complexes and the bottleneck distance for barcodes 
%This result was generalized in the papers.

%\subsection{distance function }

%\subsection{Constructing numerical measure from $\text{Cross-Barcode}_{*}(P,Q)$, MTop-Div score}

\subsection{The Manifold Topology Divergence (MTop-Div)}
The bound from eq.(\ref{eq:hausd}) and the equality $\text{Cross-Barcode}_{*} (P,P)=\varnothing$ imply that the closeness of $\text{Cross-Barcode}_{*}(P,Q)$ to the empty set is a natural measure of discrepancy between $\cal{P}$ and $\cal{Q}$   .
%Various numerical characteristics capture this discrepancy. 
Each $\text{Cross-Barcode}_{i}(P,Q)$ is a list of intervals describing the persistent homology $H_i$. 
To measure the closeness to the empty set, one can use 
segments' statistics: sum of lengths, sum of squared lengths, number of segments, the maximal length (H$_i$ max) or specific quantile.
%; alternatively -  a histogram of relative living times \cite{khrulkov2018geometry} and its distance to the histogram of the empty barcode.
We assume that various characteristics of different $H_i$ could be useful in various cases, but the cross-barcodes for $H_0$ and $H_1$ can be calculated relatively fast.

Our \textbf{MTop-Divergence}(${\cal{P}},{\cal{Q}}$) is based on the sum of lengths of segments in $\text{Cross-Barcode}_{1}(P,Q)$, see section \ref{sec:Alg} for details. 
The sum of lengths of segments in $\text{Cross-Barcode}_{1}(P,Q)$ has an interesting interpretation via the Earth Mover's Distance. Namely, it is easy to prove (see Appendix 
\ref{app:h1-rlt-proof}) that EM-Distance between the Relative Living Time histogram for $\text{Cross-Barcode}_{1}(P,Q)$ and the histogram of the empty barcode, multiplied by the parameter $\alpha_{max}$ from the definition of RLT, see e.g. \cite{khrulkov2018geometry},
%for $\alpha_{max}$ bigger than all the ``death'' times in $H_1$, 
coincides with the sum of lengths of segments in $H_1$. This ensures the standard good stability properties of this quantity.

%It is easy to prove that 
%Sum of lengths in H1/ $alpha_{max}$ is the EM-Distance from distribution concentrated at $0$ for RLT (for $alpha_{max}$ bigger than maximal death value in H1)

%We consider the distributions of  several numerical characteristics on the set of samples from $\cal{P}$, $\cal{Q}$:
%Number of segments in $H^i$
%Longest segment in $H^i$
%Sum of lengths of all segments in $H^i$
%Sum of squares of segments in $H^i$
%1)sum of lengths of segments in H1
%2)Number of segments in H1
%3)"$L_2$-Loss"=sum of squares of lengths of all segments in H1
%4)Max length of barcodes in H1, the 97,7\%(mean+2sigma), 84,1\%(mean+sigma), mean, square mean
%5)RLT-score (wasserstein ili L2)

Our metrics can be applied in two settings: to a pair of distributions  $\cal{P}_{\mathrm{data}}$, $\cal{Q}_{\mathrm{model}}$, in which case we denote our score MTop-Div(D,M), and to a pair of distributions $\cal{Q}_{\mathrm{model}}$, $\cal{P}_{\mathrm{data}}$, in which case our score is denoted MTop-Div(M,D). 
These two variants of the Cross-Barcode, and of the MTop-Divergence are  related to the concepts of precision and recall. These two variants can be analyzed separately or combined together, e.g. averaged.

%Definition

%\subsection{}Whitehead theorem on vanishing of relative homology and homotopy equivalence

\subsection{Algorithm}\label{sec:Alg}
To calculate the score that evaluates the similitude between two distributions, we employ the following algorithm. First, 
we compute $\text{Cross-Barcode}_{1} (P,Q)$ on point clouds $P,Q$
 of sizes $b_P$, $b_Q$ sampled from the two distributions $\cal{P}$, $\cal{Q}$. For this we calculate the matrices $m_P$, $m_{P,Q}$ of pairwise distances within the cloud $P$ and between clouds $P$ and $Q$. Then the algorithm constructs the  Vietoris-Rips  filtered simplicial complex from the matrix $m_{(P\cup Q)/Q}$ which is the matrix of pairwise distances in $P\cup Q$ with the pairs of points from cloud $Q$ block replaced by zeroes and with other  blocks given by $m_P$, $m_{P,Q}$. Next step is to calculate the barcode of the constructed filtered simplicial complex. This step and the previous step constructing the filtered complex from the matrix $m_{(P\cup Q)/Q}$ can be done using one of the fast scripts\footnote{\href{https://en.wikipedia.org/wiki/Persistent_homology\#Computation}{Persistent Homology Computation (wiki)}}, some of them are optimized for GPU acceleration, see e.g. \cite{zhang2020gpu,bauer2019ripser}. 
 The calculation of barcode from the filtered complex is based on the persistence algorithm 
 bringing the filtered complex to its "canonical form"  (\cite{B94}).
 Next, sum of lengths or one of other numerical characteristcs of  $\text{Cross-Barcode}_{1} (P,Q)$ is computed.
%Using numerical tests, the H1sum characteristic was picked as the principal, however depending on the situation, other characteristics can also be interesting to compute, like H1max, RLT or  H0sum.
Then this computation is run a sufficient number of times to obtain the mean value of the picked characteristic. In our experiments we have found that for common datasets the number of times from 10 to 100 is generally sufficient.   Our method  is summarized in the Algorithms \ref{alg:crossB} and \ref{alg:mtop}.

\begin{figure}[t]
\begin{multicols}{2} \setlength{\tabcolsep}{1pt}
\begin{algorithm}[H]%[tb]
   \caption{$\text{Cross-Barcode}_{i}(P,Q)$}
   \label{alg:crossB}
\begin{algorithmic}
  % \hskip-0.5em
   \STATE\hskip-1em {\bfseries Input:} $m[P,P]$, $m[P,Q]$ : matrices of pairwise distances within point cloud $P$, and between point clouds $P$ and $Q$
   %\STATE {\bfseries Input:} : matrix of pairwise distances 
   \STATE\hskip-1em {\bfseries Require:} $\text{VR}(M)$: function computing  filtered complex from pairwise distances matrix \nolinebreak$M$
   \STATE\hskip-1em {\bfseries Require:} $\text{B}(C,i)$: function computing persistence intervals of filtered complex $C$ in dimension $i$
   \STATE $b_Q\gets$ number of columns in matrix $m[P,Q]$ 
   \STATE $m[Q,Q]\gets\text{zeroes}(b_Q,b_Q)$ %zero $b_Q\times b_Q$ matrix 
   \STATE $M\gets \begin{pmatrix}
  m[P,P] & m[P,Q]\\\
 m[P,Q] & m[Q,Q]
 \end{pmatrix}$
   \STATE $\text{Cross-Barcode}_{i}\gets \text{B}(\text{VR}(M),i)$
   \STATE \hskip-1em{\bfseries Return:} list of intervals $\textbf{Cross-Barcode}_{i} (P,Q)$ representing "births" and "deaths" of topological discrepancies 
\end{algorithmic}
\end{algorithm}
\columnbreak
\begin{algorithm}[H]%[tb]
   \caption{MTop-Divergence($\cal{P},\cal{Q}$), see section \ref{sec:Alg} for details, default suggested values: $b_{\cal{P}}=1000$, $b_{\cal{Q}}=10000$, $n=100$ }
   \label{alg:mtop}
\begin{algorithmic}
   \STATE \hskip-1em {\bfseries Input:} $X_{\cal{P}}$, $X_{\cal{Q}}$:  $N_{\cal{P}}\times D$, $N_{\cal{Q}}\times D$ arrays representing datasets
   \FOR{$j=1$ {\bfseries to} $n$}
     \STATE $P_j\gets$ random choice($X_{\cal{P}}$,$b_{\cal{P}}$)
     \STATE $Q_j\gets$ random choice($X_{\cal{Q}}$,$b_{\cal{Q}}$)
     \STATE ${\cal{B}}_j\gets$ list of intervals Cross-Barcode$_{1}(P_j,Q_j)$ calculated by Algorithm\ref{alg:crossB}
     \STATE $mtd_j\gets$ sum of lengths of all intervals in ${\cal{B}}_j$
   \ENDFOR
   \STATE $\text{MTop-Divergence}({\cal{P}},{\cal{Q}})\gets\text{mean}(mtd)$
   %\STATE $r\gets$ mean distance to the closest neighbor in a sample of 1000 points from $\cal{P}_{\mathrm{data}}$
   %\STATE $\text{Normalized MTop-Divergence}({\cal{P}},{\cal{Q}})\gets\text{mean}(mtd)/ r$
   \STATE \hskip-1em{\bfseries Return:} number \textbf{MTop-Divergence}($\cal{P},\cal{Q}$)
   %, and \textbf{Normalized }\textbf{MTop-Divergence}(${\cal{P}},{\cal{Q}}$) 
   representing discrepancy between the distributions $\cal{P},\cal{Q}$
\end{algorithmic}
\end{algorithm}
\end{multicols}
\vskip-2em
\end{figure}

\textbf{Complexity.}
The Algorithm \ref{alg:crossB} requires computation of the two matrices of pairwise distances  $m[P,P]$, $m[P,Q]$   for a pair of samples  $P\in {\mathbb{R}}^{b_{\cal{P}}\times D}$, $Q\in {\mathbb{R}}^{b_{\cal{Q}}\times D}$ 
involving  $O(b_{\cal{P}}^2D)$ and
${{O}}(b_{\cal{P}}b_{\cal{Q}}D)$ operations. After that, the complexity of the computation of  barcode does not depend on the dimension $D$ of the data. Generally the persistence algorithm is at worst cubic in the number of simplices involved. In practice,  the boundary matrix  is sparse in our case and  thanks also to the GPU optimization, the computation of cross-barcode takes similar time as in the previous step on datasets of big dimensionality. Since only the discrepancies in manifold topology are calculated, the results are quite robust and a relatively low number of iterations is needed to obtain accurate results.  Since the algorithm scales linearly with $D$ it can be applied to the most recent datasets with $D$ up to $10^7$. For  example, for $D=3.15\times10^6$, and batch sizes $b_{\cal{P}}=10^3, b_{\cal{Q}}=10^4$, on  NVIDIA TITAN RTX %and 40Intel(R) Xeon(R) CPU 2.20GHz,%
the time for GPU accelerated calculation of pairwise distances was 15 seconds, and GPU-accelerated calculation of  $\text{Cross-Barcode}_{1}(P,Q)$ took 30 seconds. 

%One computation of $\text{Cross-Barcode}_{i}(P,Q)$ by Algorthm \ref{alg:crossB} takes ??? on ??(colab). 
%for $i=1$  requires at worst $O(b_{\cal{P}}b^2_{\cal{Q}})$ operations.  

\section{Experiments}\label{sec:Exper}

We examine the ability of MTop-Div to measure quality of generative models trained on various datasets. Firstly, we illustrate the behaviour of MTop-Div on simple synthetic datasets (rings (Fig.\ref{fig:rings_H0_sum}), disks (Fig. \ref{fig:two_circles})).
Secondly, we show that MTop-Div is superior to the GScore, another topology-based GAN quality measure. We carry out experiments with a mixture of Gaussians, MNIST, CIFAR10, X-ray images, FFHQ. The performance of MTop-Div is on par with FID. For images, MTop-Div is always calculated in pixel space without utilizing pre-trained networks.
Thirdly, we apply MTop-Div to GANs from non-image domains: 3D shapes and time-series, where FID is not applicable. We show that MTop-Div agrees with domain-specific measures, such as JSD, MMD, Coverage, discriminative score, but MTop-Div better captures evolution of generated data manifold during training \footnote{we  additionally calculated IMD  \cite{tsitsulin2019shape} for the pairs of point clouds from our experiments, see Appendix \ref{app:imd}.}.

\subsection{Simple synthetic datasets in 2D}  
\label{sec:synthetic}
%Cloud pairs can be divided into three types : coincident, intersecting, and non-intersecting (example on Figure  \ref{fig:two_circles}.).

%disks, spheres rings 
%demonstrating sensibility to the relative position of two clouds as opposed to Geometry-score
%demonstrate that barcode and all scores vanish as number of points increases if the two clouds are %sampled from the same distribution 

%Let's push the centers of the circles apart and detect changes barcodes. 
%\begin{figure}[h!]
%\centering 
% \includegraphics[width = 0.6\columnwidth]{img/disks/disks_H0_max.png}
% \caption{Barcodes H0 max len and Gscore metrics for disks with radius 1 per distance between centers of distributions}
 %\label{fig:disks_H0_max}
%\end{figure}

%\begin{figure}[h!]
%\centering 
% \includegraphics[width = 0.6\columnwidth]{img/disks/disks_H1_sum.png}
% \caption{Barcodes H1 sum len for disks with radius 1 per distance between centers of distributions}
% \label{fig:disks_H1_sum}
%\end{figure}

%Similar results are obtained for ring distributions.

%\begin{figure}[h!]
%\centering 
% \includegraphics[width=0.2\columnwidth]{img/rings/coinciding_rings.png}

% \includegraphics[width=0.4\columnwidth]{img/rings/disjoint_rings.png}
 % \caption{Tree types of clouds : coinciding, overlapping, non-intersecting}
 %\label{fig:two_rings}
%\end{figure}

\begin{figure}[h!]
\centering 
 \includegraphics[width=0.21\textwidth]{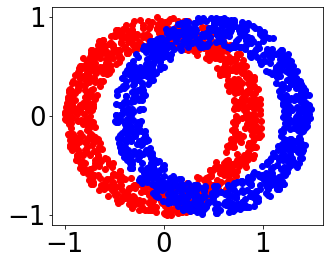}
\includegraphics[width=0.26\textwidth]{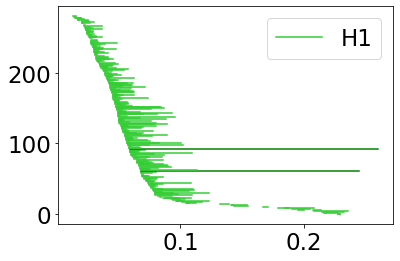}
 \includegraphics[width = 0.23\textwidth]{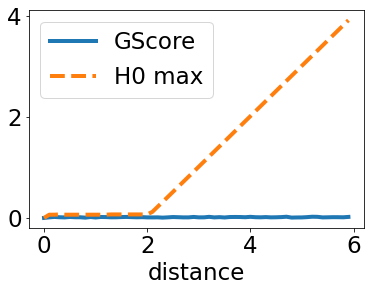}
 \includegraphics[width = 0.23\textwidth]{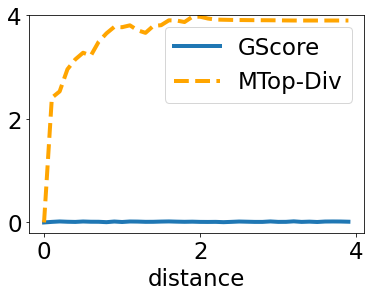}
 \caption{MTop-Div and H0 max compared with GScore, for two ring clouds of 1000 points, as function of $d=$distance between ring centers, the $\text{Cross-Barcode}_{1} (P,Q)$ is shown at $d=0.5$  }
\label{fig:rings_H0_sum}
\end{figure} As illustrated on Fig. \ref{fig:rings_H0_sum} the GScore does not respond to shifts of the distributions' relative position. 

\begin{wrapfigure}{r}{0.45\textwidth}
\centering 
\vskip-0.4in
\includegraphics[width=0.45\textwidth]{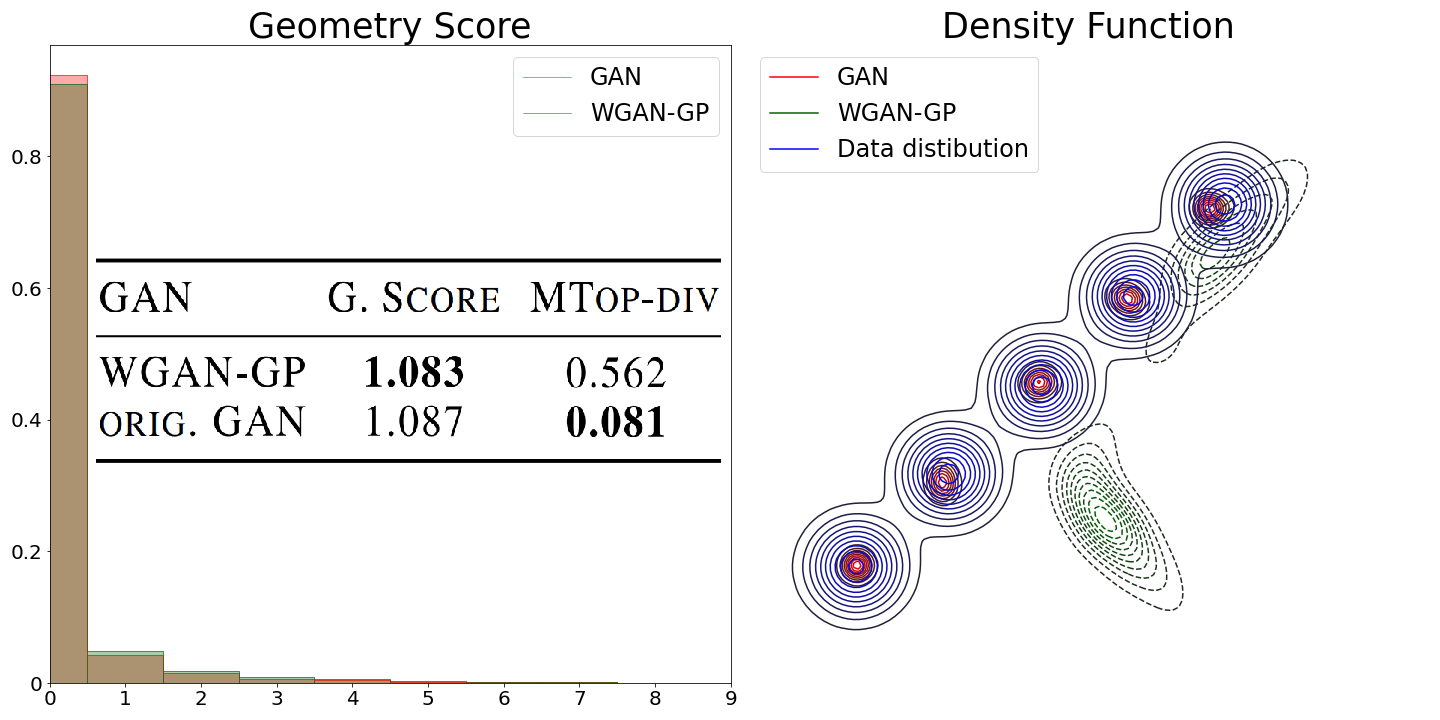}
\caption{Difficulty of the Geometry Score to detect the mode dropping.}
\label{fig:moddrop}
\vskip-0.01in
\end{wrapfigure}

\subsection{Mode-Droping on Gaussians}
\label{sec:gaussians}
One of common tasks to assess GAN's performance is measuring it's ability to uncover the variety of modes in the data distribution. A common benchmark for this problem is a mixture of Gaussians, see Fig. \ref{fig:moddrop}. 
We trained two generators with very different performance: original GAN, which managed to capture all 5 modes of the distribution and WGAN-GP, which have only covered poorly two. However, the Geometry score is not sensitive to such a difference since two point clouds have the same RLT histogram. While the MTop-Div is sensitive to such a difference. 

\subsection{Digit flipping on MNIST}
\label{sec:exp-mnist}

\begin{wrapfigure}{r}{0.58\linewidth}
\vskip-0.5in
%\begin{minipage}{0.58\textwidth}
\
%\begin{figure}[t]
\centering 
%\begin{minipage}{0.59\textwidth}
\begin{tabular}{m{1.5cm} m{0.3cm} m{1.5cm} m{3cm}}
\includegraphics[width=0.1\textwidth]{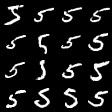}  & 
vs
& 
\includegraphics[width=0.1\textwidth]{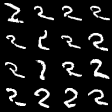}
&
\begin{tabular}{@{}c@{}}Geometry Score = 0.0 \\ MTop-Div = 6154.0 \end{tabular}
\end{tabular}
\caption{Two point clouds:``5''s from MNIST vs. vertically flipped ``5''s from MNIST (resembling rather ``2''s). The two clouds are indistinguishable for Geometry Score, while the MTop-Div is sensitive to such flip as it depends on the positions of clouds with respect to each other.}
\label{fig:fives_rotated}

\end{wrapfigure}

Figure \ref{fig:fives_rotated} shows an experiment with MNIST dataset. We compare two point clouds: ``5''s vs. vertically flipped ``5''s  (resembling rather ``2''s). These two clouds are indistinguishable for Geometry Score, while the MTop-Div is sensitive to such flip since it depends on the relative position of the two clouds.

\subsection{\mbox{Synthetic modifications of CIFAR10}}
\label{sec:cifar10}
\begin{figure*}[h!]
\centering 
 \includegraphics[width=\textwidth]{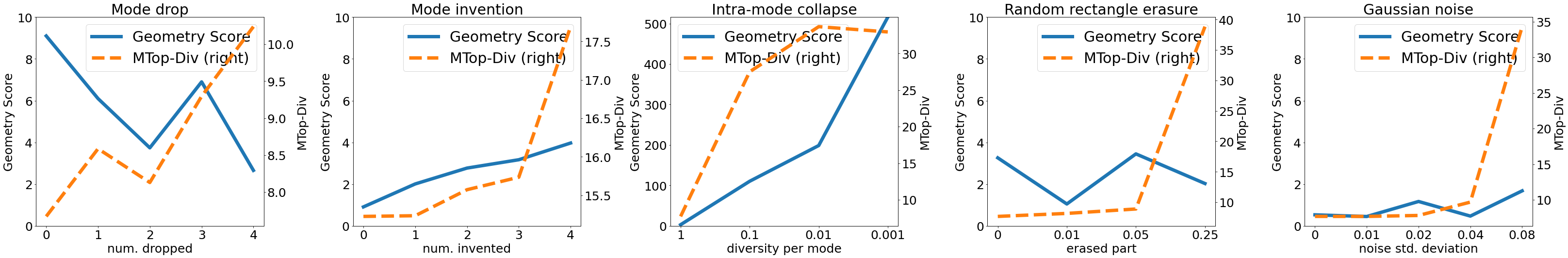}
  \caption{Experiment with modifications of CIFAR10. The disturbance level rises from zero to a maximum. Ideally, the quality score should monotonically increase with the disturbance level.}
 \label{fig:cifar10-synthetic}
\end{figure*}

We evaluate the proposed MTop-Div(D,M) using a benchmark with the controllable disturbance level. We take CIFAR10 and apply the following modifications. Firstly, we emulate common issues in GAN: mode drop, mode invention and mode collapse by doing class drop, class addition and intra-class collapse (removal of objects within a class).
Secondly, we apply two disturbance types: erasure of a random rectangle and a Gaussian noise. As `real' images we use the test set from CIFAR10, as `generated' images - a subsample of the train set with applied modifications. The size of the later always equals the test set size.
%Thus, `real' and `generated' samples always have the same size which is important since some GAN evaluation measures are sensitive to dataset sizes.
Figure \ref{fig:cifar10-synthetic} shows the results. Ideally, the quality measure should monotonically increase with the disturbance level. We conclude that Geometry Score is monotone only for `mode invention' and `intra-mode collapse' while  MTop-Div(D,M) is almost monotone for all the cases.
The average Kendall-tau rank correlation between  MTop-Div(D,M) and disturbance level is 0.89, while for Geometry Score the rank correlation is only 0.36. FID performs well on this benchmark, not shown on Figure \ref{fig:cifar10-synthetic} for ease of perception. Additionally, we calculated MTop-Div for higher order Cross-Barcodes, see Appendix \ref{app:high-mtd}.

%Why not  MTop-Div(M,D)?

%Difficulty of G-score (rotation etc., channel shuffle)-text ?

%Difficulties  FID (? not done): - Grigory
%\subsection{Comparison of GAN's}

%\begin{table}[t]
%\caption{RelTop score across datasets and architectures}
%\label{tbl:archi}
%\vskip 0.15in
%\begin{center}
%\begin{small}
%\begin{sc}
%\begin{tabular}{lcccc}
%\toprule
%GAN & MNIST & F-MNIST & SVHN & CIFAR10 \\
%\midrule
%dcgan & 1.0 & 4.75  & 162.08 & 109.27\\
%WGAN  & 1.0 &  8.25 & 234.33 & 87.27  \\
%WGAN-GP  & 0.7 & 15.86 & 712.57 & 88.83  \\
%lsgan & 0.7 & 19.75 & 1011.53 & 35.42 \\
%rel gan & 0.7 & 19.75 & 1011.53 & 35.42 \\
%sngan & 0.7 & 19.75 & 1011.53 & 35.42 \\
%\bottomrule
%\end{tabular}
%\end{sc}
%\end{small}
%\end{center}
%\vskip -0.1in
%\end{table}

%\begin{wrapfigure}{r}{0.4\textwidth}
%\centering 
%\vskip-0.4in
%\includegraphics[width=0.4\textwidth]{img/stylegan_ffhq.png}
%\caption{Comparison of the quality measures, FID vs MTop-Divergence, on StyleGAN, StyleGAN2 trained on FFHQ with different truncation levels. MTop-Div(M,D) is measured  in the pixels space, FID is measured  using the embedding in the last hidden layer of the Inception neural network. MTop-Divergence(M,D) is monotonically increasing in good correlation with FID.}
%\vskip-0.4in
%\label{fig:stylegan_ffhq}
%\end{wrapfigure}

\begin{wraptable}{r}{0.708\linewidth}
\vskip-0.19in
%\begin{minipage}{0.58\textwidth}
%\begin{figure}[t]
\centering 
\caption{MTop-Div is consistent with FID for model selection of GAN's trained on various datasets.}
    \begin{tabular}{ccccc}
        \toprule 
        Dataset & \multicolumn{2}{c}{FID} & \multicolumn{2}{c}{MTop-Div(D,M)}\\
        \midrule
        & WGAN & WGAN-GP & WGAN & WGAN-GP\\
        \cmidrule{2-5}
        CIFAR10      & \textbf{154.6} & 399.2 & \textbf{353.1} & 1637.4 \\
        SVHN         & \textbf{101.6} & 154.7 & \textbf{332.0} & 963.2 \\
        MNIST        & 31.8  & \textbf{22.0} & 2042.8 & \textbf{1526.1} \\
        FashionMNIST & 52.9  & \textbf{35.1} & 919.6 & \textbf{660.4} \\
        \bottomrule
    \end{tabular} 
    \vskip-0.1in
    \label{tab:wgan_vs_wgangp}
\end{wraptable}

\subsection{GAN model selection}
\label{sec:gan-model-selection}
We trained WGAN and WGAN-GP on various datasets: CIFAR10, SVHN, MNIST, FashionMNIST and evaluated their quality, see Table \ref{tab:wgan_vs_wgangp}. Experimental data show that the ranking between WGAN and WGAN-GP is consistent for FID and MTop-Div.

\subsection{Experiments with StyleGAN}

We evaluated the performance of StyleGAN \cite{karras2019stylebased} and StyleGAN2 \cite{karras2020analyzing} generators trained on the FFHQ dataset\footnote{\url{https://github.com/NVlabs/ffhq-dataset} (CC-BY 2.0 License)}. We generated $20\times10^3$ samples with two truncation levels: $\psi=0.7, 1.0$ and compared them with $20\times10^3$ samples from FFHQ. The truncation trick is known to improve average image quality but decrease image variance. Figure \ref{fig:stylegan_ffhq} show the results (see also Table \ref{tbl:ffhq-stylegan} in Appendix for more data). 
Thus, the ranking via MTop-Div(M, D) is consistent with FID.
We also tried to calculate Geometry Score but found that it takes a prohibitively long time.

%\subsection{Computational considerations}
%For experiments we used NVIDIA TITAN RTX and 40 Intel(R) Xeon(R) CPU 2.20GHz. The largest point cloud in our experiments had dimensionality $D=196\times10^3$. For this point cloud with  batch sizes $b_{\cal{P}}=10^3, b_{\cal{Q}}=10^4$ the majority of time took parallel calculation of pairwise distances - 18 min., while GPU-accelerated barcodes calculation took only 0.5 min. 

\subsection{\mbox{Chest X-rays generation for COVID-19 detection}}
\begin{figure}[t]
\centering 
%\begin{subfigure}[b]{0.64\textwidth}
\begin{minipage}{0.64\textwidth}
\includegraphics[width=\textwidth]{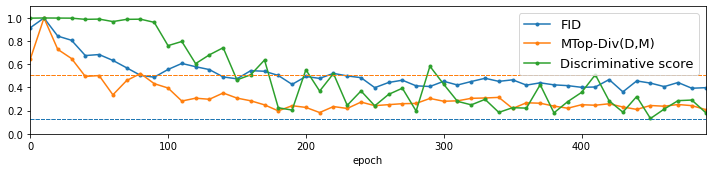}\\
\includegraphics[width=\textwidth]{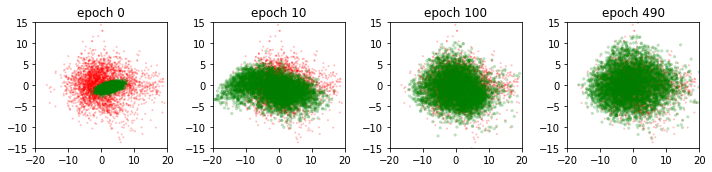}\\
\caption{Training process of GAN applied to chest X-ray data. \textbf{Top}: normalized quality measures FID, MTop-Div, Disc. score vs. epoch. Each measure is divided by it's maximum value: max. FID = 304, max MTop-Div = 21, max. Disc.score = 0.5. Lower is better. Dashed horizontal lines depict comparison of real COVID-positive and COVID-neg. chest X-rays. \textbf{Bottom}: PCA projections of real objects (red) and generated objects (green).}
\label{fig:covidgan}
%\end{subfigure}
\end{minipage}
\hfill
%\begin{subfigure}[b]{0.34\textwidth}
\begin{minipage}{0.34\textwidth}
\centering 
\includegraphics[width=0.85\textwidth]{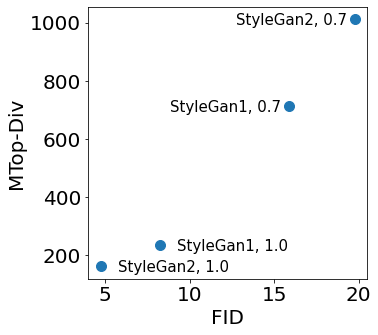}
\caption{Comparison of the quality measures, FID vs MTop-Div, on StyleGAN, StyleGAN2 trained on FFHQ with different truncation levels. %MTop-Div(M,D) is measured in the pixels space, FID is measured  using the embedding in the last hidden layer of the Inception neural network.
MTop-Div(M,D) is monotonically increasing in good correlation with FID.}
\label{fig:stylegan_ffhq}
%\end{subfigure}
\end{minipage}
\end{figure}

%Medical imaging is an actual application of GANs. 
Waheen et al. \cite{waheed2020covidgan} described how to apply GANs to improve COVID-19 detection from chest X-ray images.
Following \cite{waheed2020covidgan}, we trained an ACGAN \cite{odena2017conditional} on a dataset 
consisting of chest X-rays of COVID-19 positive and 
healthy patients \footnote{We used the ACGAN implementation \url{https://github.com/clvrai/ACGAN-PyTorch}, (MIT License) and chest X-ray data was from \url{https://www.kaggle.com/tawsifurrahman/covid19-radiography-database} (Kaggle Data License)}. Next, we studied the training process of ACGAN. Every 10'th epoch we evaluated the performance of ACGAN by comparing real and generated COVID-19 positive chest X-ray images. That is, we calculated FID, MTop-Div(D,M) and a baseline measure - discriminative score \footnote{Discriminative score equals accuracy minus 0.5. MTop-Div better correlates with the discriminative score than FID: 0.75 vs. 0.66.} of a CNN trained to distinguish real vs. generated data.
%Experimental data shows that FID is around 300 which is unrealistically high (compare it with GANs trained on FFHQ where FID $\sim10-20$), moreover it changes very slowly.
The MTop-Div agrees with FID and the discriminative score.
%At same time, MTop-Div better correlates with the discriminative score than FID : 0.7 vs. 0.6.
PCA projections show that generated data approximates real data well. Figure \ref{fig:cxr-examples} in Appendix presents real and generated images.

Additionally, we compared real COVID-positive and COVID-negative chest X-ray images, see horizontal dashed lines at Fig. \ref{fig:covidgan}. Counterintuitively, for FID real COVID-positive images are closer to real COVID-negative ones than to generated COVID-positive images; probably because FID is overly sensitive to textures. At the same time, evaluation by MTop-Div is consistent.

\subsection{3D shapes generation}

\begin{figure}[t]
\centering 
%\begin{subfigure}[t]{0.49\textwidth}
\begin{minipage}{0.49\textwidth}
\includegraphics[width=\textwidth]{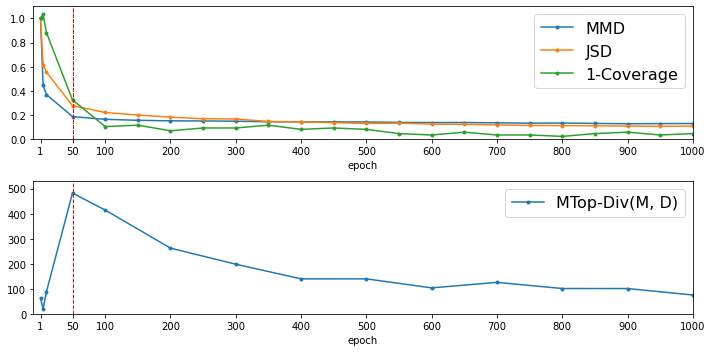}\\
\includegraphics[width=\textwidth]{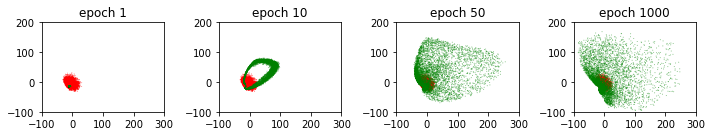}\\
\caption{Training process of GAN applied to 3D shapes. \textbf{Top, middle}: quality measures MMD, JSD, 1-Coverage, MTop-Div vs. epoch. Each quality measure is normalized, that is, divided by its value at the first epoch. Lower is better. \textbf{Bottom}: PCA projection of real objects (red) and generated objects (green). \textbf{Vertical red line} (epoch 50) depicts the moment, when the manifold of generated objects ``explodes'' and becomes much more diverse.}
\label{fig:3dgan}
\end{minipage}
%\end{subfigure}
\hfill
\begin{minipage}{0.49\textwidth}
%\begin{subfigure}[t]{0.49\textwidth}
\centering
\vskip-0.3in
\includegraphics[width=\textwidth]{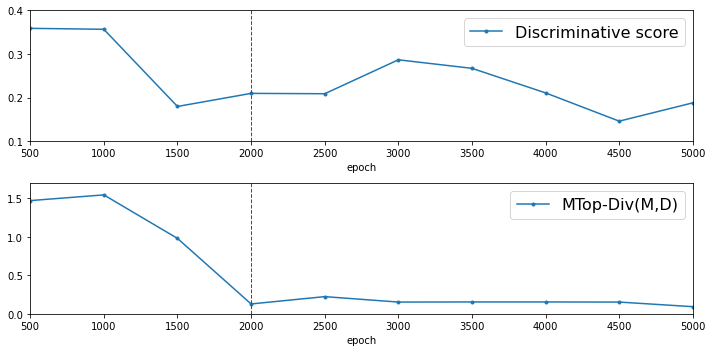}\\
\includegraphics[width=\textwidth]{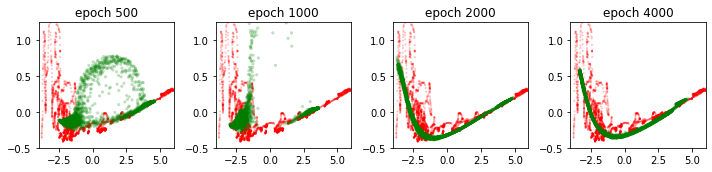}\\
\caption{Training dynamics of TimeGAN applied to market stock data. \textbf{Top}: discriminative score vs. epoch, MTop-Div vs. epoch. Lower is better. \textbf{Bottom}: PCA projection of real time-series (red) and generated time-series (green). \textbf{Vertical red line} (epoch 2000) depicts the moment when manifolds of real and generated objects become close.}
\label{fig:timegan}
%\end
\end{minipage}

\end{figure}

We use the proposed MTop-Div score to analyze the training process of the GAN applied to 3D shapes, represented by 3D point clouds \cite{achlioptas2018learning}.
%The trainset consist of 3D shapes represented by 3D point clouds.
For training, we used 6778 objects of the ``chair'' class from ShapeNet \cite{chang2015shapenet}. We trained GAN for 1000 epochs and tracked the following standard quality measures: Minimum Matching Distance (MMD), Coverage, and Jensen-Shannon Divergence (JSD) between marginal distributions.
To understand the training process in more details, we computed PCA decomposition of real and generated objects (Fig. \ref{fig:3dgan}, bottom). For computing PCA, each object (3D point cloud) was represented by a vector of point frequencies attached to the 3-dimensional $28^3$ grid. 
Figure \ref{fig:3dgan}, top, shows that conventional metrics (MMD, JSD, Coverage) doesn't represent the training process adequately.
While these measures steadily improve, the set of generated objects dramatically changes. At epoch 50, the set of generated objects (green) ``explodes'' and becomes much more diverse, covering a much larger space than real objects (red). Conventional quality measures (MMD, JSD, Coverage) ignore this shift while MTop-Div has a peak at this point. Next, we evaluated the final quality of GAN by training a classifier to distinguish real and generated object. A simple MLP with 3 hidden layers showed accuracy 98\%, indicating that the GAN poorly approximates the manifold of real objects. This result is consistent with MTop-Div: at epoch 1000 it is even larger than at epoch 1.

\subsection{Time series generation}

Next, we analyze training dynamics of TimeGAN \cite{yoon2019time} tailored to multivariate time-series generation. We followed the experimental protocol from \cite{yoon2019time} and used the daily historical market stocks data from 2004 to 2019, including as features the volume, high, low, opening, closing, and adjusted closing prices. 
The baseline evaluation measure is calculated via a classifer (RNN) trained to distinguish real and generated time-series.
Particularly, the \textit{discriminative score} equals to the accuracy of such a classifer minus 0.5.
Fig. \ref{fig:timegan}, top, shows the results. We conclude that the behaviour of MTop-Div is consistent with the discriminative score: both of them decrease during training. To illustrate training in more details we did PCA projections of real and generated time-series by flattening the time dimension (Fig. \ref{fig:timegan}, bottom). At 2000-th epoch, the point clouds of real (red) and generated (green) time-series became close, which is captured by a drop of MTop-Div score. At the same time, discriminative score is not sensitive enough to this phenomena.

\section{Conclusions}\label{sec:concl}
We have proposed a tool, $\text{Cross-Barcode}_{*}(P,Q)$, which records multiscale topology discrepancies between two data manifolds approximated by point clouds.
Based on the $\text{Cross-Barcode}_{*}(P,Q)$, we have introduced the Manifold Topology Divergence and have applied it to evaluate the performance of GANs in various domains: 2D images, 3D shapes, time-series. We have shown that the MTop-Div correlates well with domain-specific measures and can be used for model selection. Also, it provides insights about the evolution of generated data manifold during training and can be used for early stopping. The MTop-Div score is domain agnostic and does not rely on pre-trained networks. We have compared the MTop-Div against 7 established evaluation methods: FID, discriminative score, MMD, JSD, 1-coverage, IMD, and Geometry score and found that MTop-Div outperforms many of them and captures well subtle differences in data manifolds. Our methodology permits to overcome the known TDA scalability issues and to carry out the MTop-Div calculations on the most recent datasets such as FFHQ, with the size of up to $10^5$ and the dimension of up to $10^7$.  

% to assess the performance of deep generative models on various datasets: MNIST, CIFAR10, FFHQ.
% We have found a better performance of the MTop-Divergence compared with GScore on various tasks, including detection of mode-dropping.
% We have demonstrated that the MTop-Divergence accurately detects  degrees of intra-mode collapse, mode invention, mode-dropping, and gaussian noise. Our algorithm scales well (essentially linearly) with the increase of the dimensionality of data. Our methodology can be applied to datasets of different sizes, including the ones on which the recent GANs in the visual domain are trained. 

% We apply the MTop-Div to evaluate performance of GANs in various domains: 2D images, 3D shapes, time-series. We show that the MTop-Div correlates well with domain-specific measures and can be used for model selection. Also it provides insights about evolution of generated data manifold during training;

%   We have essentially overcame the known TDA scalability issues and in particular have carried out the Mtop-Div calculations on most recent datasets such as FFHQ, with dimensions $D$ up . 
%%%%%%%%%%%%%%%%%%%%%%%%%%%%%%%%%%%%%%%%%%%%%%%%%%%%%%%%%%%%
\subsection*{Acknowledgements}
The problem statement was developed in the framework of Skoltech-MIT NGP program. The work of Serguei Barannikov and Evgeny Burnaev was supported by Ministry of Science and Higher Education grant No. 075-10-2021-068. Authors are thankful to Alexei Artemov for help with the 3D shapes experiment.
%\clearpage

\bibliography{references}
\bibliographystyle{plain}

\clearpage

\appendix

\section{Simplicial Complexes, Cycles, Barcodes}

\subsection{Background}
The  simplicial complex is a combinatorial data that can be thought of as a higher-dimensional generalization of a graph.
Simplicial complex $S$ is a collection of  $k-$simplices,  which are finite $(k+1)-$elements subsets in a given set $V$, for $k\geq 0$.  The collection of simplices $S$ must satisfy the condition that for each $\sigma \in S$,  $\sigma' \subset \sigma$ implies $\sigma'\in S$. A simplicial complex consisting only of $0-$ and $1-$simplices is a graph. 

Let $C_k(S)$ denotes the vector space over a field $F$ whose basis elements are $k-$simplices from $S$ with a choice of ordering of vertices up to an even permutation. 
%An arbitrary reordering $\gamma$ of the simplex vertices changes the basis element in $C_k(S)$  by $(-1)^{\bar\gamma}$. 
In calculations it is most convenient to put $F=\mathbb{Z}_2$. The boundary linear operator $\partial_k:C_k(S)\to C_{k-1}(S)$ is defined on $\sigma=\{x_0,\ldots,x_{k}\}$ as
\begin{equation*}
    \partial_k \sigma=\sum_{j=0}^k (-1)^j\{x_0,\ldots,x_{j-1},x_{j+1},\ldots,x_{k}\}.
\end{equation*}
The $k-$th \textbf{homology} group $H_k(S)$ is defined as the vector space  $\ker\partial_k/\operatorname{im}\partial_{k+1}$. The elements $c\in \ker\partial_k$ are called \textbf{cycles}. The elements $\tilde{c}\in\operatorname{im}\partial_{k+1}$ are called boundaries. The general elements $c'\in C_k(S)$ are called chains.  The elements of  $H_k(S)$ represent various $k-$dimensional topological features in $S$. A basis in $H_k(S)$ corresponds to a set of basic topological features. 

Filtration on simplicial complex is defined as a family of simplicial complexes $S_\alpha$ with nested collections of simplices: for $\alpha_1 < \alpha_2$ all simplices of $S_{\alpha_1}$ are also in $S_{\alpha_2}$. In practical examples the indexes $\alpha$  run through a discrete set $\alpha_1<\ldots<\alpha_{\max}$. 

The inclusions $S_{\alpha}\subseteq S_{\beta}$ induce naturally the maps on the homology groups $H_k(S_{\alpha})\to H_k(S_{\beta})$.
The evolution of the cycles through the nested family of simplicial complexes $S_{\alpha}$ is described by the barcodes. The  persistent homology principal theorem  \cite{B94,zomorodian2005computing,zomorodian2001computing} states that for each dimension there exists a choice of a set of basic topological features across all $S_{\alpha}$ so that each feature appears in $H_k(S_{\alpha})$ at specific time $\alpha=b_j$ and disappears at specific time $\alpha=d_j$. The $H_i$ barcode of the filtered simplicial complex is the record of these times represented as the collection of segments $[b_j,d_j]$.  The barcodes are defined and calculated through bringing the set of matrices of the boundary operators $\partial_k$ to the ''Canonical Form'' by a change of the basis in $C_k$ preserving the nested family $S_\alpha$ \cite{B94,barannikov2021canonical}.

Let $(\Gamma,m)$ be a weighted graph with distance-like weights, where $m$ is the symmetric matrix of the weights attached to the edges of the graph $\Gamma$. The Vietoris-Rips filtered simplicial complex of the weighted graph $ R_\alpha(\Gamma,m)$, is defined as the nested collection of simplices:
\begin{equation*}
   R_\alpha(\Gamma,m)=\left\{\{x_0,\ldots,x_k\}, x_i\in\operatorname{Vert}(\Gamma)\Vert m(x_j,x_l)\leq \alpha \right\}
\end{equation*}
where $\operatorname{Vert}(\Gamma)$ is the set of vertices of the graph $\Gamma$. 
Even though such weighted graphs do not always come from a set of points in metric space, barcodes of weighted graphs have been successfully applied in many situations (networks, fmri, medical data, graph's classification etc). 

\subsection{Simplices, describing discrepancies between the two manifolds}
\begin{figure}[h]
\centering 
 \includegraphics[width=0.20\columnwidth]{clouds/ALPHA02_.png}
 \includegraphics[width=0.20\columnwidth]{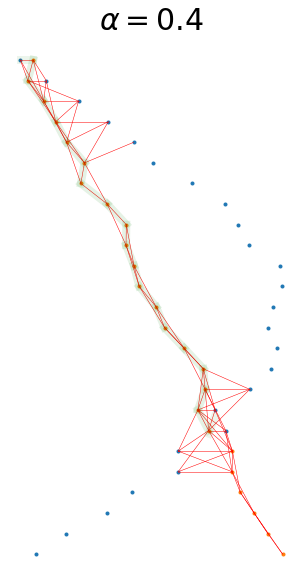}
 \includegraphics[width=0.20\columnwidth]{clouds/ALPHA06.png}
 \includegraphics[width=0.20\columnwidth]{clouds/ALPHA08_.png}
   \caption{We are adding edges between $P-$points(orange) and $Q-$points(blue) and between pairs of $P-$points consecutively. The edges are sorted by their length, and are added at the threshold $\alpha\geq 0$ corresponding to the length of the edge. Here at the thresholds $\alpha=0.2, 0.4, 0.6, 0.8$ edges with length less than $\alpha$ were added. For ease of perception the simultaneously added triangles and higher simplices,  as well  as the added at $\alpha=0$ all simplices between $Q-$points,  are not shown explicitly here. Notice how the $1-$cycle,  shown with green, with endpoints in $Q-$cloud is born at $\alpha=0.4$. It survives at $\alpha=0.6$ and it is killed at $\alpha=0.8$. }
 \label{fig:edges2}
\end{figure}

\begin{figure}[h]
\centering 
 \includegraphics[width=0.18\columnwidth]{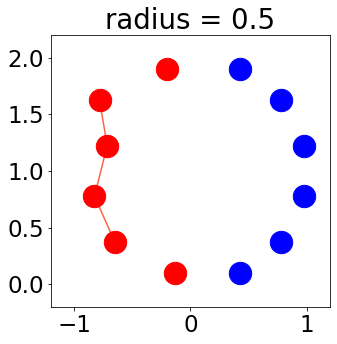}
 \includegraphics[width=0.18\columnwidth]{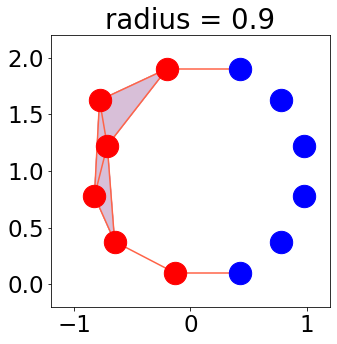}
 \includegraphics[width=0.18\columnwidth]{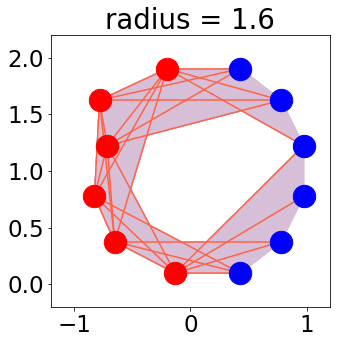}
  \includegraphics[width=0.18\columnwidth]{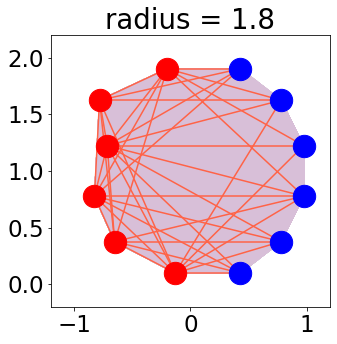}
  \includegraphics[width=0.24\columnwidth]{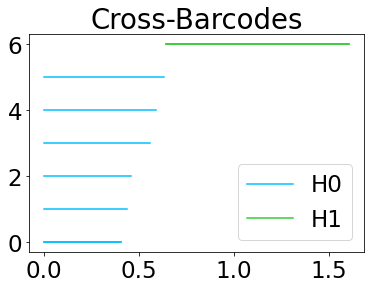}
  \caption{The process of adding the simplices between the $P-$cloud(red) and $Q-$cloud(blue) and within the $P-$cloud. Here we show the consecutive adding of edges together with simultaneous adding of triangles. All the edges and simplices within $Q-$cloud are assumed added at $\alpha=0$ and are not shown here for perception's ease. Notice the $1-$cycle born between $\alpha=0.5$ and $0.9$, it corresponds to the green segment in the shown Cross-Barcode}
 \label{fig:visualization_barcodes}
\end{figure}

Here we gather more details on the construction of sets of simplicies that describe discrepancies between two point clouds $P$ and $Q$ sampled from the two distributions $\cal{P}$, $\cal{Q}$. 
As we have described in section 2, our basic methodology is to add consecutively the edges between $P-$points and  $Q-$points and between pairs of $P-$points. All edges between $Q-$points are added simultaneously at the beginning at the threshold $\alpha=0$ . The $PP$ and $PQ$ edges are sorted by their length, and are added at the threshold $\alpha\geq 0$ corresponding to the length of the edge. This process is visualized in more details on Figures \ref{fig:edges2} and \ref{fig:visualization_barcodes}.  The triangles are added at the threshold at which the last of its three edges are added. The $3-$ and higher $k-$simplices are added similarly at the threshold corresponding to the adding of the last of their edges.  The added triangles and higher dimensional simplices are not shown explicitely on Figure \ref{fig:edges} for ease of perception, as they can be restored from their edges. As all simplices within the $Q-$cloud are added at the very beginning at $\alpha=0$, the corresponding cycles formed by the $Q-$cloud simplices are immediately killed at $\alpha=0$ and do not contribute to the Cross-Barcode.

The constructed set of simplices is naturally a simplicial complex, since for any added $k-$simplex, we have added also all its $(k-1)-$faces obtained by deletion of one of vertices. The threshold $\alpha$ defines the filtration on the obtained simplicial complex, since the simplices added at smaller threshold $\alpha_1$ are contained in the set of simplices added at any bigger threshold $\alpha_2>\alpha_1$.

With adding more edges, the cycles start to appear. In our case, a cycle is essentially a collection of simplices whose boundary is allowed to be nonzero if the boundary consists of simplices with vertices from $Q$. For example, a $1-$ cycle in our case is a path consisting of added edges, that can start and end in $Q-$cloud and that passes through $P-$points.  This is because any such collection can be completed to a collection with zero boundary since any cycle from $Q-$cloud is a boundary of a sum of added at $\alpha=0$ simplices from $Q$.

A $1-$cycle disappears at the threshold when a set of triangles is added whose boundary coincides with the $1-$cycle plus perhaps some edges between $Q-$points. 

Notice how the $1-$cycle with endpoints in $Q-$cloud is born at $\alpha=0.4$ on Figure  \ref{fig:edges2}, shown with green. It survives at $\alpha=0.6$ and it is killed at $\alpha=0.8$. 
The process of adding longer edges can be visually assimilated to the building of a "spider's web" that tries to bring the cloud of red points closer to the cloud of blue points. The obstructions to this are quantified by "lifespans" of cycles, they correspond to the lengths of segments in the barcode.  See e.g., Figure \ref{fig:visualization_barcodes} where a $1-$cycle is born between $\alpha=0.5$ and $0.9$, it then corresponds to the green segment in the Cross-Barcode. 

%At the initial moment, $\alpha=0$, the simplicial complex $R_\alpha(\Gamma_{P\cup Q},m_{(P\cup Q)/Q})$ has trivial homology $H_k$ for all $k>0$ since it consists of all simplices formed by $Q-$points. The dimension of the $0-$th homology  equals at $\alpha=0$  to the number of $P-$points, since no edge between them or between a $P-$point and a $Q-$point is added at the beginning.
%\begin{proposition}
%The chain groups of the 
%\end{proposition}

\begin{remark}
To characterise the situation of two data point clouds one of which is a subcloud of the other $S\subset C$,  it can be tempting to start seeking a "relative homology" analog of the standard (single) point cloud persistent homology. The reader should be warned that the common in the literature "relative persistent homology" concept and its variants, 
 i.e. the persistent homology of the decreasing sequence of factor-complexes of a fixed complex: $K\to\dots K/K_i\to K/K_{i+1}\to\dots K/K$, 
is irrelevant in the present context. In contrast, our methodology, in particular, does not involve factor-complexes construction, which is generally computationally prohibitive. 
The point is that the basic concept of filtered complex contains naturally its own relative analogue via the appropriate use of various filtrations.  
\end{remark}

%\begin{figure}[th]
%\centering 
% \includegraphics[width=0.45\columnwidth]{img/edges/alpha02.png}
 %\includegraphics[width=0.45\columnwidth]{img/edges/alpha03.png}
% \includegraphics[width=0.45\columnwidth]{clouds/ALPHA04.png}
% \includegraphics[width=0.45\columnwidth]{img/edges/alpha06.png}
% \includegraphics[width=0.45\columnwidth]{img/edges/alpha08.png}
%   \caption{We are adding edges between $P-$points(orange) and $Q-$points(blue) and between pairs of $P-$points consecutively. At the thresholds $\alpha=0.2, 0.4, 0.6, 0.8$ edges with length less than $\alpha$ are added. For ease the simultaneous adding of triangles is not shown explicitly here. }
% \label{fig:edges}
%\end{figure}
\subsection{Sub-manifolds and bars in {Cross-Barcode}$_{*}(P,Q)$}
It is natural to start analyzing the closeness of the data point cloud $P$ to the data point cloud $Q$ by looking at the matrix of the $PQ$ pairwise distances. If there are many points from $P$ such that their distance to their closest point from $Q$ is relatively big then the clouds $P$ and $Q$ are not close. However, in applications, it is important to distinguish the different situations here. The first case is when all these remote from Q points  are close to each other. Then this remote from $Q$ cluster of $P$ points represents a single topological feature distinguishing cloud $P$ from $Q$. Another case is when the remote from $Q$ points form several clusters so that each such remote from $Q$ cluster represents a separate topological feature. The long bars in the zero-dimensional Cross-Barcode record the lifespans on the distances’ scale of these remote from $Q$ clusters of $P-$points.

In practice it also happens more often that it is not possible to distinguish a separate cluster of $P$ points which are all remote from $Q$. Rather, there are some $P-$points inside the same $P-$cluster that are close to $Q$ and other $P-$points from the same $P-$cluster which are further away from $Q$, as on Fig.\ref{fig:edges}. This situation is captured and quantified by the higher dimensional topological features distinguishing cloud $P$ from $Q$. Intuitively such an $i-$dimensional topological feature represents an $i-$dimensional $P-$cloud's sub-manifold whose boundary is close to the $Q-$cloud, but whose interior $P-$points are remote from the $Q-$cloud, like the green polygonal chain on Fig.\ref{fig:edges2} at $\alpha=0.4$. Such features are constructed in the algorithm using the distance matrix combinatorics from $(i+1)-$tuples of $P-$points or $P$ \& $Q-$ points. The distances within each of these tuples are less or equal to the feature's appearance, or birth, threshold. The disappearance, or death, of such a feature calculated by the algorithm corresponds approximately to the scale at which the feature becomes indistinguishable from the $Q-$cloud. The $i-$dimensional {Cross-Barcode}$_{i}(P,Q)$, $i\geq 1$, is the set of segments (bars) recording the birth and the death thresholds of such topological features.

\subsection{$\text{Cross-Barcode}_{*}(P,Q)$ as obstructions to assigning $P$ points to distribution $\cal{Q}$}
Geometrically, the lowest dimensional  $\text{Cross-Barcode}_0(P,Q)$ is the record of relative hierarchical clustering of the following form. For a given threshold $r$, let us consider all points of the point cloud $Q$ plus the points of the cloud $P$ lying at a distance less than $r$ from a point of $Q$ as belonging to the single $Q-$cluster.
It is natural to form simultaneously other clusters based on the threshold $r$, with the rule that  if the distance between two points of $P$ is less than threshold $r$ then they belong to the same cluster. 
When the threshold $r$ is increased, two or more clusters can collide. And the threshold, at which this happens, corresponds precisely to the ``death'' time of one or more of the colliding clusters. At the end, for very large $r$ only the unique $Q-$cluster survives. Then $\text{Cross-Barcode}_0(P,Q)$ records the survival times for this relative clustering.  

%Joining two parts of the same cluster. 

\begin{figure}[h!]
\centering 
 \includegraphics[width=0.22\columnwidth]{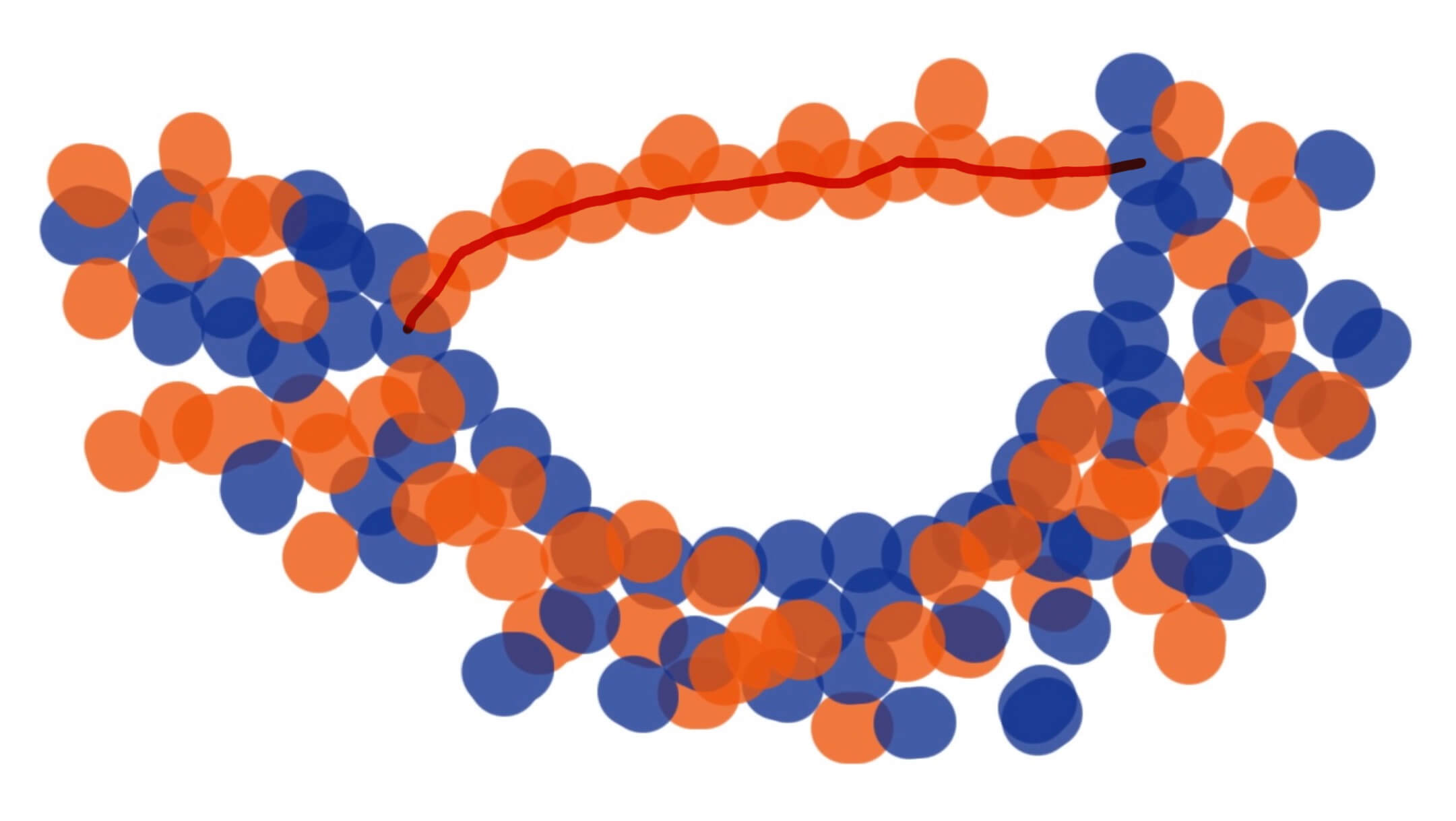} 
 \includegraphics[width=0.22\columnwidth]{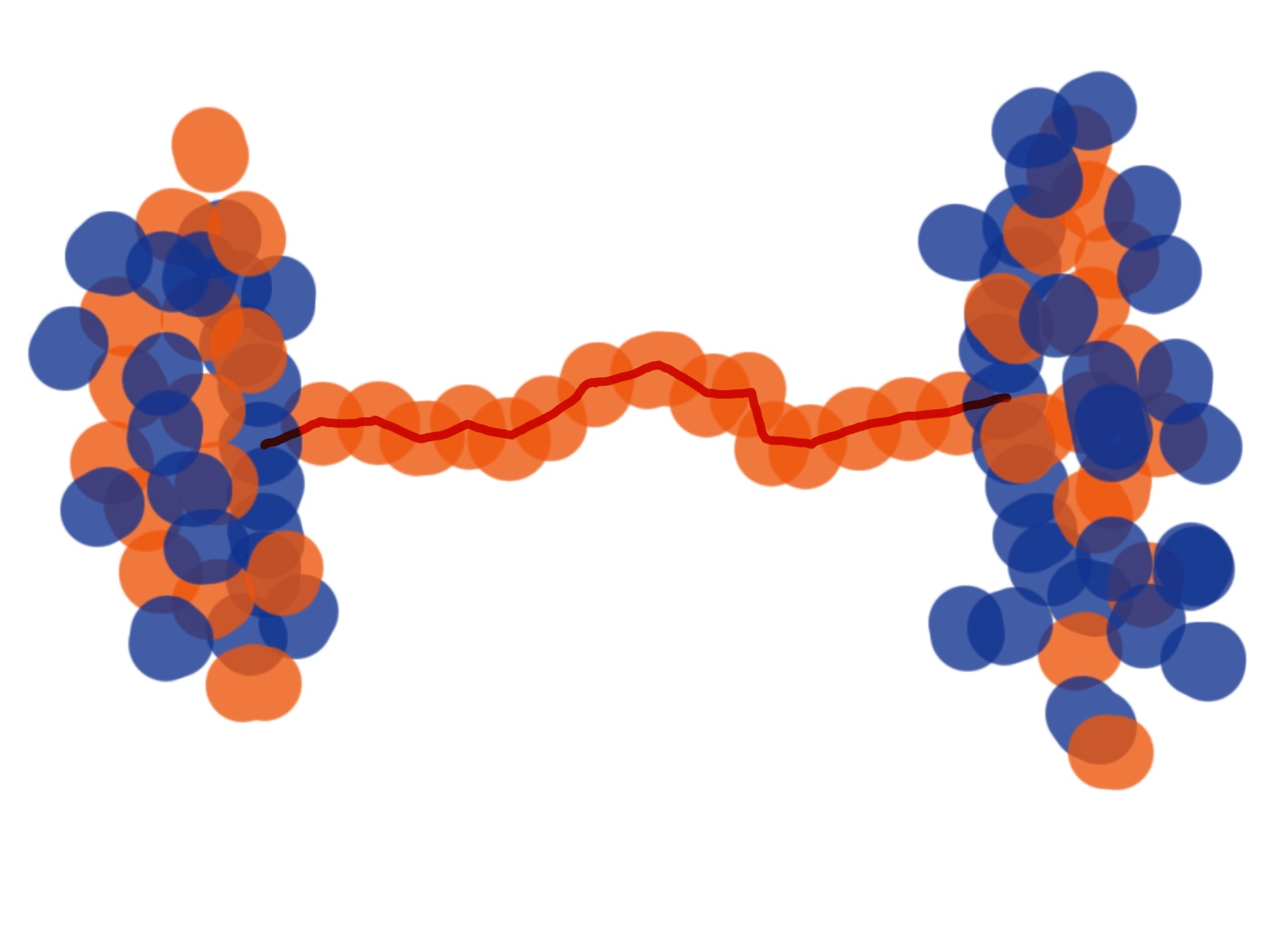} \hspace{0.8cm}
 \includegraphics[width=0.33\columnwidth]{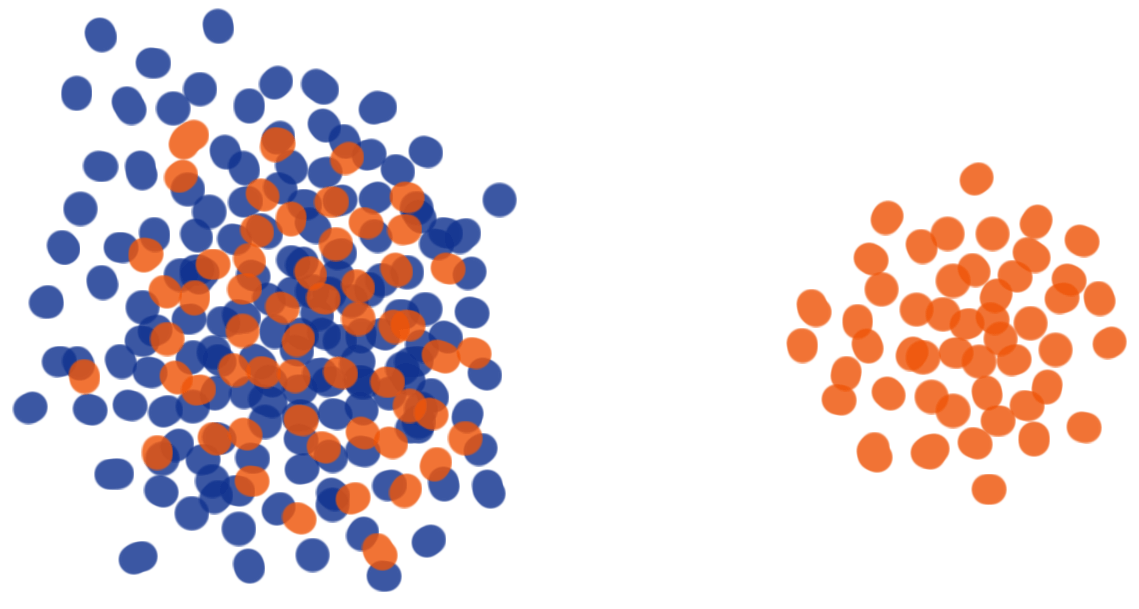}
  \caption{ Paths/membranes (red) in the void that are formed by small intersecting disks around $P$ points (orange), and are ending on $Q$ (blue), are obstacles for identification of the distribution  $\cal{P}$ with  $\cal{Q}$. These obstacles are quantified by  $\text{Cross-Barcode}_{1} (P,Q)$. Separate clusters are the obstacles  quantified by $\text{Cross-Barcode}_{0} (P,Q)$.}
 \label{fig:obst}
\end{figure}
Notice that in situations, like, for example, in Figure \ref{fig:obst}, it is difficult to attribute confidently certain points of $P$ to the same distribution as the point cloud $Q$ even when they belong to the ``big'' $Q-$cluster at a small threshold $r$, because of the nontrivial topology. Such ``membranes'' of $P-$points in void space,  are obstacles for assigning  points from $P$ to distribution $\cal{Q}$.  These  obstacles  are quantified by the segments from the higher barcodes $\text{Cross-Barcode}_{\geq 1} (P,Q)$. The bigger the length of the associated segment in the barcode, the further the membrane passes away from $Q$. 

\subsection{More simple synthetic datasets in 2D}

\begin{figure}[h!]
\centering 
 \includegraphics[width=0.3\columnwidth]{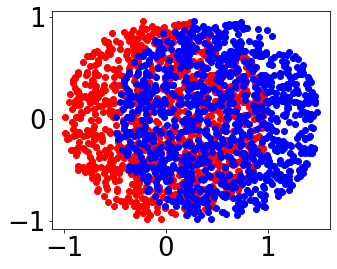}
 \includegraphics[width=0.29\columnwidth]{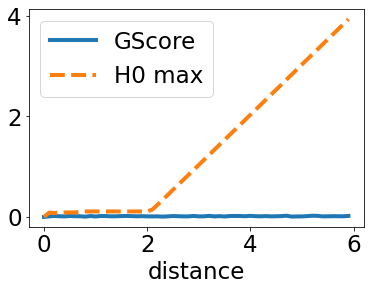}
  \includegraphics[width=0.29\columnwidth]{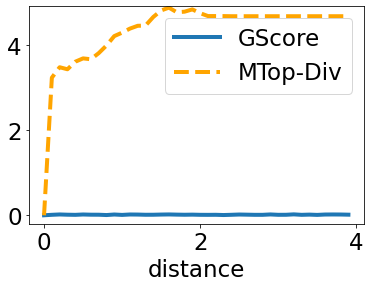}
  \caption{The first picture shows two clouds of 1000 points sampled from the uniform distributions on two different  disks of radius 1 with a distance between the centers of the disks of 0.5. The second and third pictures show the dependence of the GScore metric (the GScore is equal to zero independently of the distance between the disks), maximum of segments in $H0$ and the sum of lengths of segments H1 as a function of the distance between the centers of the disks averaged over 10 runs. The length of the maximum segment barcode in H0  grows linearly and  equals to the distance between the pair of closest points in the two distributions. }
 \label{fig:two_circles}
\end{figure}

 %We demonstrate properties of 
 %first verify  are interested in the results of counting barcodes in dimensions 0 and 1, where variables such as the total length or number of barcodes in each dimension will show important information about the similarity of clouds and their distributions.
 
% We are interested in the results of counting barcodes in dimensions 0 and 1, where variables such as the total length or %number of barcodes in each dimension will show important information about the similarity of clouds and their distributions.
%Cloud pairs can be divided into three types : coincident, intersecting, and non-intersecting (example on Figure  \ref{fig:two_circles}.).

%disks, spheres rings 
%demonstrating sensibility to the relative position of two clouds as opposed to Geometry-score
%demonstrate that barcode and all scores vanish as number of points increases if the two clouds are %sampled from the same distribution 

%Let's push the centers of the circles apart and detect changes barcodes. 
%\begin{figure}[h!]
%\centering 
% \includegraphics[width = 0.6\columnwidth]{img/disks/disks_H0_max.png}
% \caption{Barcodes H0 max len and Gscore metrics for disks with radius 1 per distance between centers of distributions}
 %\label{fig:disks_H0_max}
%\end{figure}

%\begin{figure}[h!]
%\centering 
% \includegraphics[width = 0.6\columnwidth]{img/disks/disks_H1_sum.png}
% \caption{Barcodes H1 sum len for disks with radius 1 per distance between centers of distributions}
% \label{fig:disks_H1_sum}
%\end{figure}
As illustrated on Figures \ref{fig:rings_H0_sum},\ref{fig:two_circles} the GScore is unresponsive to changes of the distributions' positions. 

\subsection{Cross-Barcode and precision-recall}

\begin{figure}[h!]
\centering 
 \includegraphics[width=0.4\textwidth]{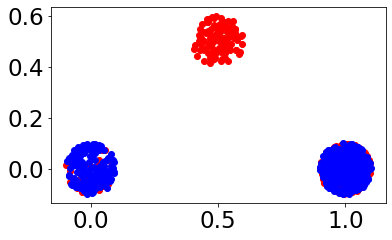}
 \includegraphics[width=0.4\textwidth]{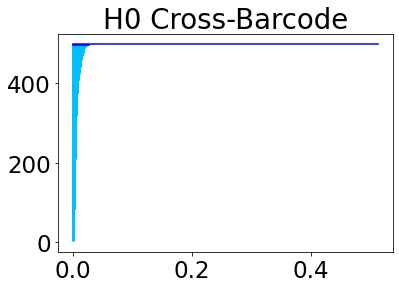}
  \caption{Mode-dropping, bad recall \& good precision,  illustrated with  clouds $P_{\text{data}}$ (red) and $Q_{\text{model}}$ (blue). The Cross-Barcode$_0(P_{\text{data}}, Q_{\text{model}})$ contains long intervals, one for each dropped mode, which measure the distance from the data's dropped mode to the closest generated mode.}
 \label{fig:mode-dropping}
\end{figure}

The Cross-Barcode captures well the precision vs. recall aspects of the point cloud's approximations, contrary to FID, which is known to mix the two aspects. For example, in the case of mode-dropping, bad recall but good precision, the Cross-Barcode$_0(P_{\text{data}},Q_{\text{model}})$ contains the long intervals, one for each dropped mode, which measure the distance from the data's dropped mode to the closest generated mode. 
The mode-dropping case (bad recall, good precision) is illustrated on Figure \ref{fig:mode-dropping}.
\begin{figure}[h!]
\centering 
 \includegraphics[width=0.4\textwidth]{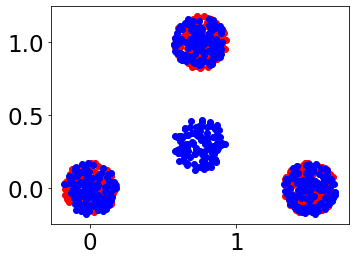}
 \includegraphics[width=0.4\textwidth]{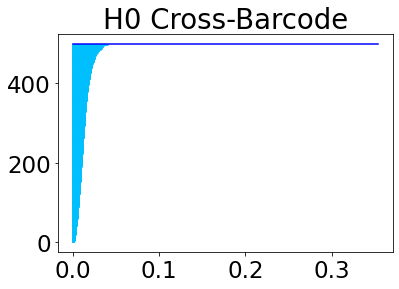}
  \caption{Mode-invention, good recall \& bad precision, illustrated with  clouds $P_{\text{data}}$ (red) and $Q_{\text{model}}$ (blue). The  Cross-Barcode$_0(Q_{\text{model}},P_{\text{data}})$ contains long intervals, one for each invented mode, which measure the distance from the model's invented mode to the closest data's mode.}
 \label{fig:mode-invent}
\end{figure}

Analogously, in the case of mode-invention, with good recall but bad precision, the Cross-Barcode$_0(Q_{\text{model}},P_{\text{data}})$ contains long intervals, one for each invented mode, which measure the distance from the model's invented mode to the closest data's mode. 

The mode-invention (good recall, bad precision) case is illustrated on Figure \ref{fig:mode-invent}.

\section{Cross-Barcode Properties.}\label{sec:appCrossProp}

\subsection{The Cross-Barcode's norm and the Hausdorf distance.}

The  {Bottleneck distance} \cite{efrat2001geometry}, also known as Wasserstein$-\infty$ distance $\mathbb{W}_{\infty}$, defines the natural norm on the Cross-Barcodes:
\begin{equation*}
      \left\Vert \text{Cross-Barcode}_{i} \left(P,Q\right)\right\Vert_B=\max_{[b_j,d_j]\in\text{Cross-Barcode}_{i}}(d_j-b_j).
  \end{equation*}
  
The Hausdorf distance measures how far are two subsets $P,Q$ of a metric space from each other. The Hausdorf distance is the greatest of all the distances from a point in one set to the closest point in the other set:
\begin{equation}
    d_{\mathrm H}(P,Q) =\max \left\{\,\sup_{x \in P} d(x,Q),\, \sup_{y \in Q} d(y,P) \,\right\}.
\end{equation}
\textbf{Proposition 1.}  \emph{The norm of  $\text{Cross-Barcode}_{i} (P,Q)$, $i\geq 0$, is bounded from above by the Hausdorff distance} 
  \begin{equation}\label{eq:hausdA}
      \left\Vert \text{Cross-Barcode}_{i} \left(P,Q\right)\right\Vert_B\leq d_H(P,Q).
  \end{equation}
\begin{proof}
Let $c\in R_{\alpha_0}(\Gamma_{P\cup Q},m_{(P\cup Q)/Q})$ be an $i-$dimensional cycle appearing in the filtered complex at $\alpha=\alpha_0$. Let us construct a simplicial chain that kills $c$. Let $\sigma=\{x_1,\ldots,x_{i+1}\}$ be one of the simplices from $c$.  Let $q_j$ denote the closest point in $Q$ to the vertex $x_j$. The prism $\{x_1,q_1,\ldots,x_{i+1},q_{i+1}\}$ can be decomposed into $(i+1)$ simplices $p_k(\sigma)=\{x_1,x_2,\ldots,x_{k-1},q_{k},\ldots,q_{i+1}\}$, $1\leq k\leq i+1$. The boundary of the prism consists of the two simplices $\sigma$,  $q(\sigma)=\{q_1,\ldots,q_{i+1}\}$, and of the $(i+1)$ similar prisms corresponding to the  the boundary simplices of $\sigma$. If $c=\sum_n a_n\sigma^n$ then 
\begin{equation*}
    c=\partial(\sum_n a_n \sum_k p_k(\sigma^n)) + \sum_n a_n q(\sigma_n)
\end{equation*}
For any $k,j$, $d(x_j,x_k)\leq \alpha_0$ since $c$ is born at $\alpha_0$. Therefore
\begin{equation*}
    d(x_j,q_k)\leq d(x_j,x_k)+d(x_k,q_k)\leq \alpha_0 + \sup_{x \in P} d(x,Q).
\end{equation*}
Therefore all simplices  $p_k(\sigma^n))$ appear no later than at $(\alpha_0 + \sup_{x \in P} d(x,Q))$ in the filtered complex. 
All vertices of the simplices $q(\sigma_n)$ are from $Q$. It follows that the lifespan of the cycle $c$ is no bigger than $\sup_{x \in P} d(x,Q))$
\end{proof}

\begin{wrapfigure}{r}{0.56\textwidth}
\centering 
%\vskip-0.5in
 \includegraphics[width=0.56\textwidth]{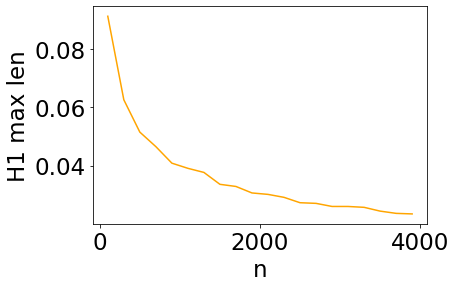}
  \caption{Diminishing of the length of maximal segment in $H_1$ with increase of the number of sampled points for point clouds of the equal size $n$ sampled from the same uniform distribution on the 2D disk of radius $1$.}
 \label{fig:H1maxpern}
\vskip-0.1in
\end{wrapfigure}

To illustrate the proposition 1 we have verified empirically the diminishing of $\text{Cross-Barcode}_{*}(Q_1,Q_2)$ when  number of points in $Q_1$, $Q_2$ goes to $+\infty$ and  $Q_1$, $Q_2$ are sampled from the same uniform distribution on the 2D disk of radius 1.
The maximal length of segments in $H1$ as function of number of points in the clouds of the same size is shown  in Figure \ref{fig:H1maxpern}. 

\subsection{MTop-Div and the Cross-Barcode's Relative Living Times (RLT) }
\label{app:h1-rlt-proof}
The Cross-Barcode for a given homology $H_i$ is a list of birth-death pairs (segments)
$$
\text{Cross-Barcode}_{i}(P,Q) = \{[b_j, d_j]\}_{j=1}^n
$$
Relative Living Times is a discrete distribution $RLT(k)$ over non-negative integers $k\in\{0, 1, \ldots, +\infty\}$.
For a given $\alpha_{max}>0$, $RLT(k)$ is a fraction of ``time'', that is, parts of horizontal axis $\tau \in [0, \alpha_{max}]$, such that exactly $k$ segments $[b_i, d_i]$ include $\tau$.

For equal point clouds, Cross-Barcode$_i(P,P)$ = $\varnothing$ and the corresponding RLT is the discrete distribution concentrated at zero. Let us denote by $O_0$ such discrete distribution corresponding to the empty set. 
A natural measure of closeness of the distribution RLT to the distribution $O_0$ is the earth-mover's distance (EMD), also called Wasserstein-1 distance.

\textbf{Proposition}. Let for all $d_i \le \alpha_{max}$, then 
\begin{equation*}
  \text{MTop-Div}(P, Q) = \alpha_{max}\mathrm{EMD}(RLT(k), O_0).  
\end{equation*}

\begin{proof} By the definition of EMD 
$$
\mathrm{EMD}(RLT, O_0) = \sum_{k=1}^{+\infty} k \times RLT(k).
$$
Let's use all the distinct $b_i, d_i$ to split $[0, \alpha_{max}]$ to disjoint segments $s_j$:
$$
[0, \alpha_{max}] = \bigsqcup_{j} s_j.
$$
Each $s_j$ is included in $K(j)$ segments $[b_i, d_i]$ from the Cross-Barcode$_i(P,Q)$.
Thus, 
$$
RLT(k) = \frac{1}{\alpha_{max}}\sum_{j : K(j) = k} |s_j|.
$$
%the total length of all the segment from the Cross-Barcode$_i(P,Q)$
At the same time:
\begin{align*}
\text{MTop-Div}(P, Q) & = \sum_i (d_i - b_i) 
 = \sum_j K(j) |s_j| = \sum_{k=1}^{+\infty} \sum_{j : K(j) = k} K(j) |s_j| \\
 & = \sum_{k=1}^{+\infty} \alpha_{max}\times k\times RLT(k)
 = \alpha_{max} \mathrm{EMD}(RLT(k), O_0).
\end{align*}
%We conclude that the proposed MTop-Div$(P, Q)$ equals EMD between RLT and zero-concentrated distribution $P_0$ up to a multiplicative constant.
\end{proof}
%For comparison of scores for different datasets, the Mtop-Div along with other characteristics measured in units of length, can be normalized by the mean distance to the closest neighbor in ${P}$.

\section{Hyperparameters, Software used, and Experiments' Details} \label{sec:hyperp}

We have made experiments in various settings and on the following datasets:
\begin{itemize}
 % \item on simple synthetic 2D datasets,  pairs of disks and  pairs of rings.  
  \item on a set of gaussians in 2D in comparison with distributions generated by  GAN and WGAN.  
  \item \textbf{MNIST} We have observed that GScore is not sensitive to the flip of the cloud of ``fives", while our score MTop-Divergence is sensitive to such flip since it depends on the positions of clouds with respect to each other
  \item \textbf{CIFAR10} We have evaluated our MTop-Div(D,M) using a benchmark with the controllable disturbance level. We have observed  that Geometry Score is monotone only for ‘mode invention’ and ‘intra-mode collapse’ while MTop-Div(D,M) is almost monotone for all the cases.
  \item \textbf{FFHQ} We have evaluated the quality of distributions generated by StyleGAN and StyleGAN-2, without truncation and with $\psi=0.7$ truncation.  We observed that the ranking via MTop-Div is consistent with FID
  \item \textbf{ShapeNet}\footnote{The dataset is free for non-commercial purposes.} We have studied the training dynamics of the GAN trained on 3D shapes. We observed that MTop-Div is consistent with domain specific measures (JSD, MMD, Coverage) and that MTop-Div better describes the evolution of the point cloud of generated objects during epochs;
  \item \textbf{Stock data} We have studied the training dynamics of TimeGAN \footnote{\url{https://github.com/jsyoon0823/TimeGAN}} applied to market stock data. We observed that MTop-Div is consistent with the discriminative score but better captures the evolution of point cloud of generated objects during epochs;
  \item \textbf{Chest X-ray images} We have studied the training dynamics of ACGAN applied to chest X-ray images.  We observed that MTop-Div is more consistent with the discriminative score than FID;
\end{itemize}  

For computation of FID we used Pytorch-FID\footnote{ \url{https://github.com/mseitzer/pytorch-fid}, (Apache Licence 2.0)}.
For computation of Geometry Score we used the original code\footnote{\url{https://github.com/KhrulkovV/geometry-score}} patched to supported multi-threading, otherwise it was extremely slow. The RLTs computation was averaged over 2500 trials. We calculated persistent homology via ripser++\footnote{\url{https://github.com/simonzhang00/ripser-plusplus}, (MIT License)}.

We used the following hyperparameters to compute \mbox{MTop-Div}:
\begin{itemize}
    \item MNIST: $b_{P}=10^2, b_{Q}=10^3$;
    \item Gaussians: $b_{P}=10^2, b_{Q}=10^3$;
    \item CIFAR10: $b_{P}=10^3, b_{Q}=10^4$;
    \item FFHQ: $b_{P}=10^3, b_{Q}=10^4$.
    \item ShapeNet: $b_{P}=10^2, b_{Q}=10^3$;
    \item Market stock data: $b_{P}=10^2, b_{Q}=10^3$;
    \item Chest X-ray data: $b_{P}=10^2, b_{Q}=10^3$.
\end{itemize}
MTop-Div scores were were averaged over 20 runs.

%\section{Experiments' additional details}
%\subsection{Details of the Experiment with Mixtures of Gaussians}

We compared Geometry Score and MTop-Div in the experiment with mixtures of Gaussians. Table \ref{tbl:gaussians} shows the results. We conclude that the MTop-Div is consistent with the visual quality of GAN's output while  Geometry Score fails.

\begin{figure}
\centering 
 \includegraphics[width=0.49\columnwidth]{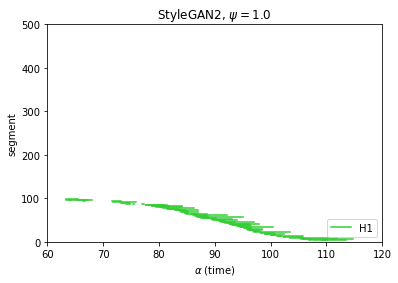}
 \includegraphics[width=0.49\columnwidth]{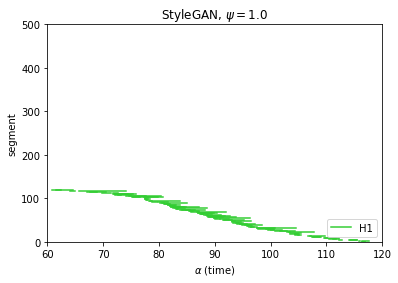}
 \includegraphics[width=0.49\columnwidth]{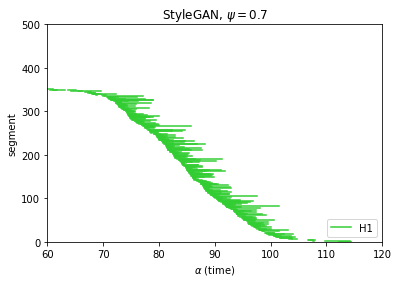}
 \includegraphics[width=0.49\columnwidth]{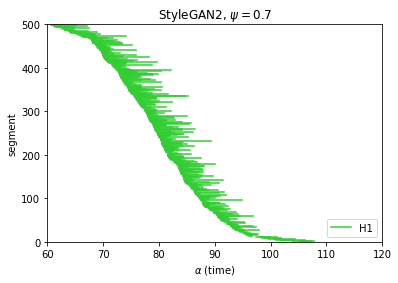}
  \caption{Cross-Barcodes for GAN's trained on FFHQ. Cross-Barcodes are shown by decrease of generator performance. For clarity, only $H_1$ barcodes are shown. The number and the total length of segments give the same ranking as the FID score}
 \label{fig:stylegan_barcodes}
\end{figure}

\begin{figure}[t!]
\centering 
 \includegraphics[width=0.49\columnwidth]{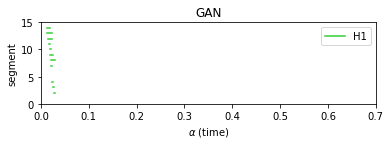}
 \includegraphics[width=0.49\columnwidth]{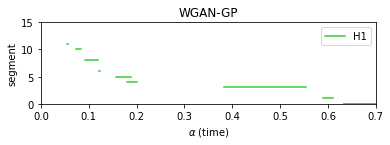}
  \caption{Cross-Barcodes for GAN's trained of mixtures of Gaussians. Cross-Barcodes are shown by decrease of generator performance. For clarity, only $H_1$ barcodes are shown.}
 \label{fig:gaussians_barcodes}
\end{figure}

\begin{figure}[t!]
\centering 
 \includegraphics[width=0.50\columnwidth]{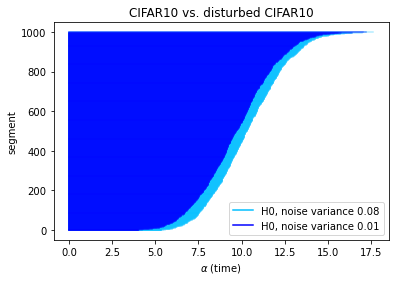}
  \caption{Cross-Barcodes for CIFAR10 vs. disturbed CIFAR10. Gaussian noise with 2 levels of variance was applied. For clarity, only $H_0$ barcodes are shown. For ease of perception of differences in Cross-Barcodes$_0$  they are shown on the same plot. The dataset with higher level of noise  is distinguished here by the longer segments in the Cross-Barcode$_0$}
 \label{fig:cifar10_barcodes}
\end{figure}

%\subsection{Examples of Cross-Barcodes from the experiments}
Figure \ref{fig:stylegan_barcodes} shows Cross-Barcodes for the experiment with StyleGAN's trained on FFHQ.
Figure \ref{fig:cifar10_barcodes} shows one of Cross-Barcodes in $H_0$ from the experiment with CIFAR10 dataset to  illustrate that the $0-$dimensional Cross-Barcode can also be applied. Figure \ref{fig:gaussians_barcodes} shows the Cross-Barcodes in $H_1$ from the experiments with GAN\footnote{https://arxiv.org/abs/1406.2661} and WGAN-GP \footnote{https://arxiv.org/abs/1704.00028} trained on mixture of Gaussians. 

%\begin{table}[tb]
%\caption{MTop-Div is consistent with FID for model selection of GAN's trained on various datasets.}
%\centering 
%    \begin{tabular}{ccccc}
%        \toprule 
%        Dataset & \multicolumn{2}{c}{FID} & \multicolumn{2}{c}{MTop-Div(D,M)}\\
%        \midrule
%        & WGAN & WGAN-GP & WGAN & WGAN-GP\\
%        \cmidrule{2-5}
%        CIFAR10      & \textbf{154.6} & 399.2 & \textbf{353.1$\pm$16.6} & 1637.4$\pm$42.7 \\
%        SVHN         & \textbf{101.6} & 154.7 & 369.6$\pm$13.5 & \textbf{348.8$\pm$12.9} \\
%        MNIST        & 31.8  & \textbf{22.0} & 2042.8$\pm$32.6 & \textbf{1526.1$\pm$30.4} \\
%        FashionMNIST & 52.9  & \textbf{35.1} & 919.6$\pm$22.7 & \textbf{660.4$\pm$18.1} \\
%        \bottomrule
%    \end{tabular} 
%    \label{tab:ext_wgan_vs_wgangp}
%\end{table}

\begin{table}
\caption{Performance measures of StyleGANs trained on FFHQ.}
\label{tbl:ffhq-stylegan}
\vskip -0.1in
\begin{center}
\begin{small}
\begin{sc}
\begin{tabular}{lcccc}
\toprule
GAN & $\psi$ & FID & MTop-Div(m,d) 
%&  MTop-Div(d,m) 
\\  
\midrule
StyleGAN2 & 1.0 &  4.75 & 162.08 
%& 109.27
\\
StyleGAN  & 1.0 &  8.25 & 234.33 
%& 87.27 
\\
StyleGAN  & 0.7 & 15.86 & 712.57 
%& 88.83
\\
StyleGAN2 & 0.7 & 19.75 & 1011.53 
%& 35.42
\\
\bottomrule
\end{tabular}
\end{sc}
\end{small}
\end{center}
\vskip -0.2in
\end{table}

\begin{table}[tb]
\caption{MTop-Div is consistent with FID for model selection of GAN's trained on various datasets.}
\centering 
    \begin{tabular}{ccccc}
        \toprule 
        Dataset & \multicolumn{2}{c}{FID} & \multicolumn{2}{c}{MTop-Div(M,D)}\\
        \midrule
        & WGAN & WGAN-GP & WGAN & WGAN-GP\\
        \cmidrule{2-5}
        CIFAR10      & \textbf{154.6} & 399.2 & \textbf{370.5$\pm$17.3} & 2408.5$\pm$27.0 \\
        SVHN         & \textbf{101.6} & 154.7 & \textbf{332.0$\pm$12.4} & 963.2$\pm$22.62 \\
        MNIST        & 31.8  & \textbf{22.0} & 2365.6$\pm$40.1 & \textbf{1474.2$\pm$29.7} \\
        FashionMNIST & 52.9  & \textbf{35.1} & 1052.6$\pm$24.8 & \textbf{872.9$\pm$21.8} \\
        \bottomrule
    \end{tabular} 
    \label{tab:ext_wgan_vs_wgangp}
\vskip 0.6in
\end{table}

%\subsection{GAN model selection}
 Table \ref{tab:ext_wgan_vs_wgangp} shows extended experimental results on GAN model selection including standard error of sample means of MTop-Div.
 
%\subsection{Experiments with StyleGAN}

%\begin{wraptable}{r}{0.5\linewidth}
%\vskip-0.3in

\begin{table}
%\begin{minipage}{0.58\textwidth}
%\begin{figure}[t]
\centering 
\caption{MTop-Div and G. Score for GAN's trained of mixtures of Gaussians.}
\label{tbl:gaussians}
%\vskip 0.15in
\begin{center}
\begin{small}
\begin{sc}
\begin{tabular}{lcccc}
\toprule
GAN & G. Score & MTop-div(m,d) &  MTop-div(d,m) & IMD\\
\midrule
WGAN-GP & \textbf{1.083} & 0.562 & 0.206 & \textbf{2.65}\\
orig. GAN  & 1.087 & \textbf{0.081} & \textbf{0.149} & 13.87 \\
\bottomrule
\end{tabular}
\end{sc}
\end{small}
\end{center}
%\vskip -0.1in
\end{table}

%\subsection{Details on chest X-ray generation}

\begin{figure}
\centering 
\includegraphics[width=0.4\columnwidth]{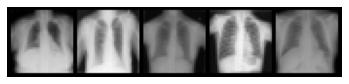}
\\
\includegraphics[width=0.4\columnwidth]{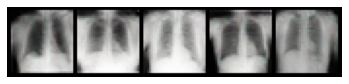}
\caption{Top: real chest X-ray images. Bottom: generated chest X-ray images.}
\label{fig:cxr-examples}
\end{figure}

Figure \ref{fig:cxr-examples} presents real and generated chest X-ray images. The generated images are of high visual quality and resembles real images.  

%\subsection{Experiments with 3D shapes}

\begin{figure}
\centering 
\includegraphics[width=0.3\columnwidth]{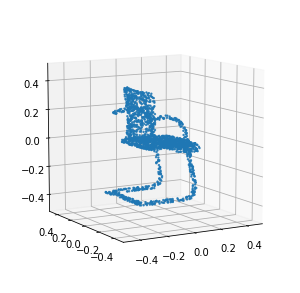}
\includegraphics[width=0.3\columnwidth]{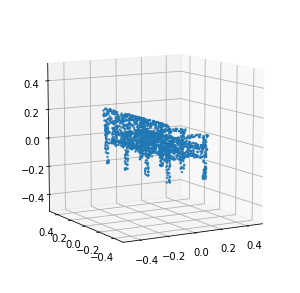}
\includegraphics[width=0.3\columnwidth]{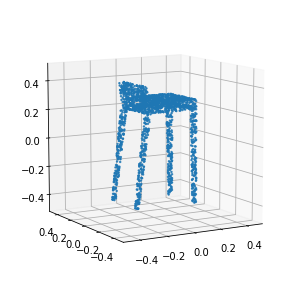}
\\
\includegraphics[width=0.3\columnwidth]{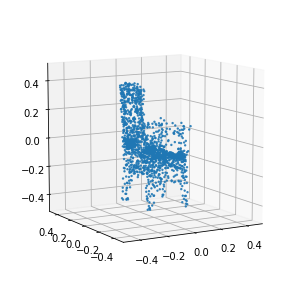}
\includegraphics[width=0.3\columnwidth]{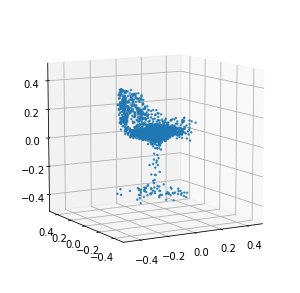}
\includegraphics[width=0.3\columnwidth]{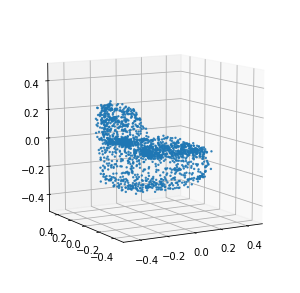}
\caption{Top: real 3D shapes. Bottom: generated 3D shapes. Generated 3D shapes are relatively blurry.}
\label{fig:3dgan-examples}
\end{figure}

Figure \ref{fig:3dgan-examples} shows real and generated 3D shapes. Generated 3D shapes (bottom row) are relatively blurry.

%%%%%%%%%%%%%%%%%%%%%%%%%%%%%%%%%%%%%%%%%%%%%%%%%%%%%%%%%%%%

\section{MTop-Div for Cross-Barcodes of higher order} \label{app:high-mtd}

We calculated MTop-Div$_k$(D,M) based on higher order Cross-Barcodes, that is, sums of segments' lengths of Cross-Barcode$_{k}$, $k>1$ were taken in Algorithm \ref{alg:mtop}. 
Then, we measured average Kendall-tau rank correlation between MTop-Div$_k$(D,M) and the disturbance level for the series of synthetic modifications of CIFAR10.
For MTop-Div$_2$(D,M) the rank correlation was 0.59, for MTop-Div$_3$(D,M): 0.45.
To make faster calculations  small batches were selected, MTop-Div$_2$(D,M): $b_{P}=100, b_{Q}=300$, MTop-Div$_3$(D,M): $b_{P}=100, b_{Q}=100$. An optimization that we describe in a future publication pre-computes the unnecessary simplices and permits faster higher degree MTop-Div computations.

\section{Comparison with the ``Intrinsic Multi-scale Distance(IMD)''}
\label{app:imd}

As proposed by a reviewer, we did additional experiments with IMD \cite{tsitsulin2019shape} applied to point clouds from our experiments. 
IMD is not sensitive to the rings shift (Section \ref{sec:synthetic}) and the digits flipping on MNIST (Section \ref{sec:exp-mnist}). For the experiment ``Mode dropping on Gaussians'' (Section \ref{sec:gaussians}), IMD incorrectly ranks poorly performing WGAN-GP (see Fig.\ref{fig:moddrop}) higher than the original GAN (Table \ref{tbl:gaussians}). For the experiments ``GAN model selection'' (Section \ref{sec:gan-model-selection}), IMD  ranks a better performing model lower in one case, while the ranking via MTop-Div is consistent with true GAN performance. For the ``Synthetic variations of CIFAR10'' experiment (Section \ref{sec:cifar10}), the average Kendall-tau correlation between IMD score and the disturbance level is 0.55, which is lower than the same measure of MTop-Div (0.89).

\end{document}